\documentclass[conference]{IEEEtran}
\usepackage[numbers, compress]{natbib}
\usepackage{bm}
\usepackage{placeins}
\usepackage{times}
\usepackage{epsfig}
\usepackage{graphicx}
\usepackage{amsmath}
\usepackage{amsthm}
\usepackage{amssymb}
\usepackage{graphicx}
\usepackage{subcaption}
\usepackage{booktabs}
\usepackage{algorithm, setspace}
\usepackage{algpseudocode, array, multirow}
\usepackage{caption}
\usepackage{xcolor}
\usepackage{extarrows}
\usepackage{booktabs}
\usepackage{rotating}
\usepackage{fixltx2e}
\usepackage{textcomp}
\usepackage{tikz}
\usepackage{soul}
\usetikzlibrary{arrows.meta, shapes.geometric, calc, shadows, matrix, positioning, angles, quotes}
\usepackage{etoolbox}
\usepackage{listofitems}
\usepackage[shortlabels]{enumitem}
\usepackage[export]{adjustbox}
\usepackage{orcidlink}

\usepackage{hyperref}
\hypersetup{colorlinks,breaklinks,urlcolor=[rgb]{0,0,0},linkcolor=[rgb]{1,0,0},citecolor=[rgb]{0,0,1}}
\usepackage[capitalize]{cleveref}

\newcommand\blfootnote[1]{%
  \begingroup
  \renewcommand\thefootnote{}\footnote{#1}%
  \addtocounter{footnote}{-1}%
  \endgroup
}

\DeclareMathOperator*{\argmax}{arg\,max}

\begin{document}
\title{HALO: Robust Out-of-Distribution Detection \\ via Joint Optimisation}

\author{

\IEEEauthorblockN{Hugo Lyons Keenan\textsuperscript{\orcidlink{0009-0006-7862-6371}}, Sarah Erfani\textsuperscript{\orcidlink{0000-0003-0885-0643}}, and Christopher Leckie\textsuperscript{\orcidlink{0000-0002-4388-0517}}}
\IEEEauthorblockA{\textit{School of Computing and Information Systems} \\
\textit{The University of Melbourne}, Victoria, Australia \\
\{lyonskeenanh, sarah.erfani, caleckie\}@unimelb.edu.au}


}

\maketitle

\begin{abstract}

Effective out-of-distribution (OOD) detection is crucial for the safe deployment of machine learning models in real-world scenarios. However, recent work has shown that OOD detection methods are vulnerable to adversarial attacks, potentially leading to critical failures in high-stakes applications. This discovery has motivated work on \textit{robust} OOD detection methods that are capable of maintaining performance under various attack settings. Prior approaches have made progress on this problem but face a number of limitations: often only exhibiting robustness to attacks on OOD data or failing to maintain strong clean performance. In this work, we adapt an existing robust classification framework, TRADES, extending it to the problem of robust OOD detection and discovering a novel objective function. Recognising the critical importance of a strong clean/robust trade-off for OOD detection, we introduce an additional loss term which boosts classification and detection performance. Our approach, called HALO (Helper-based AdversariaL OOD detection), surpasses existing methods and achieves state-of-the-art performance across a number of datasets and attack settings. Extensive experiments demonstrate an average AUROC improvement of 3.15 in clean settings and 7.07 under adversarial attacks when compared to the next best method. Furthermore, HALO exhibits resistance to transferred attacks, offers tuneable performance through hyperparameter selection, and is compatible with existing OOD detection frameworks out-of-the-box, leaving open the possibility of future performance gains. 
Code is available at: \href{https://github.com/hugo0076/HALO}{https://github.com/hugo0076/HALO}.
\end{abstract}

\begin{IEEEkeywords}
Adversarial Robustness, Out-of-distribution Detection, Image Classification
\end{IEEEkeywords}

\section{Introduction}\label{sec:intro}

\begin{figure}[t]
\centering
\includegraphics[width=0.4\textwidth]{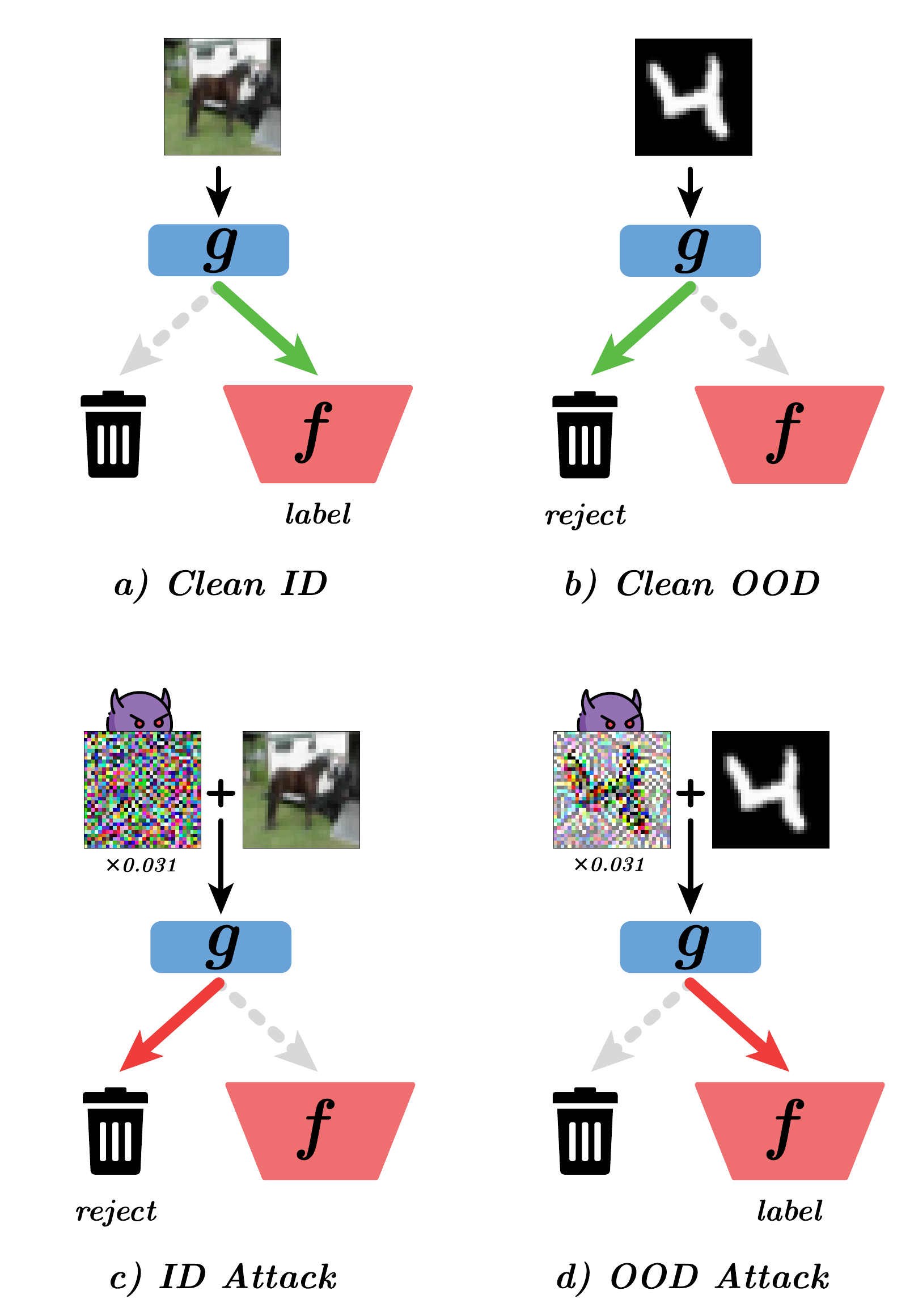}
\caption{A visualisation of two kinds of detection attack where a CIFAR-10 image (a horse) is ID and a MNIST image (a handwritten four) is OOD. $f$ is the classifier and $g$ the OOD detector. Subfigures left to right: (a) A clean ID sample is passed to the classifier and assigned a label; (b) A clean OOD sample is detected as OOD and rejected; (c) An \textbf{ID$\rightarrow$OOD attack} where the attacked ID sample is detected as OOD and erroneously rejected; and (d) An \textbf{OOD$\rightarrow$ID attack} that causes an OOD sample to evade detection and erroneously be assigned an ID label. \vspace{-1.0em}}
\label{fig:attack_vis}
\end{figure}

Neural networks have been responsible for significant improvements to image classification systems, enabling their roll-out to various real-world settings such as self-driving \cite{shen2023improved,li2019deep,bojarski2016end} and medical diagnosis \cite{yadav2019deep,goyal2020artificial,li2014medical}.\blfootnote{This work
has been accepted for publication in the IEEE Conference on Secure and
Trustworthy Machine Learning (SaTML) 2025. The final version will be
available on IEEE Xplore.} Out-of-distribution (OOD) detection is a critical capability of these systems, allowing them to identify inputs that differ significantly from the training distribution \cite{hendrycks17baseline,amodei2016concrete}. This capability is essential to the safe deployment of these systems, as encountering unfamiliar or anomalous data is inevitable in many of the high-stakes settings in which they are used. 
Concerningly, recent work has demonstrated that these systems' OOD detection performance, much like their classification performance, is vulnerable to adversarial attacks \cite{fort2022adversarial,chen2022robust}. As depicted in Figure \ref{fig:attack_vis}, these attacks come in a variety of types: targeting both in-distribution (ID) and OOD samples. By adding a small amount of carefully crafted noise to either an OOD or ID image, an adversary is able to effectively reduce or increase the system's confidence that the image is out-of-distribution, leading to a failure to separate these input types. 

\begin{figure*}[t!]
\centering
\includegraphics[width=0.95\textwidth]{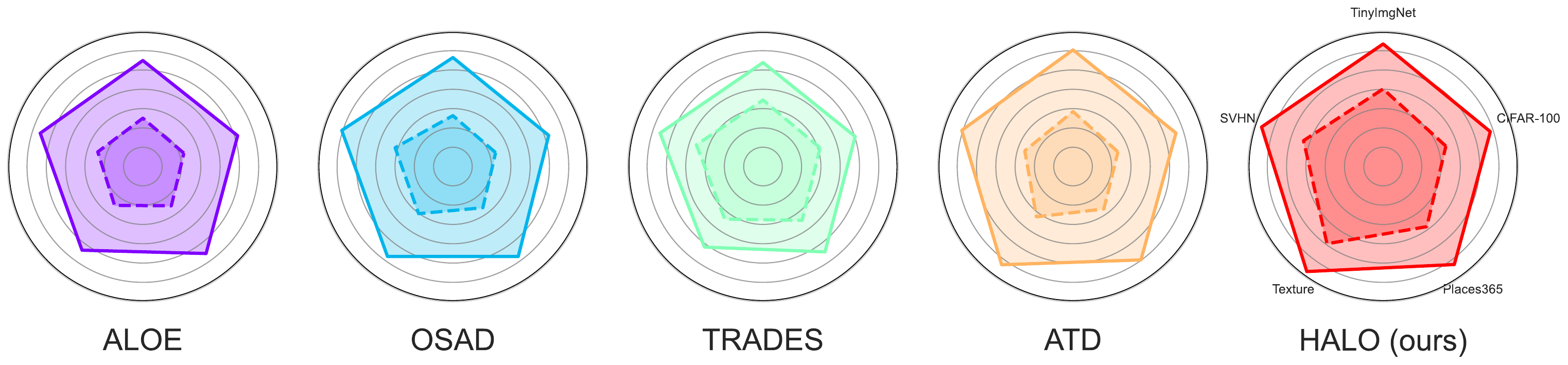}
\caption{Model detection performance (AUROC) in both standard (solid lines) and attacked (dashed lines) settings for different datasets. We compare ALOE \cite{chen2022robust}, OSAD \cite{shao2022open}, TRADES \cite{zhang2019theoretically}, ATD \cite{azizmalayeri2022your} and our method, HALO. Our method achieves the strongest clean and robust performance over a range of different datasets. \vspace{-0.75em}}
\label{fig:radar}
\end{figure*}

Such attacks have the potential to cause significant harm or failure when perpetrated in high stakes settings. Consider the case of a self-driving vehicle, for which in-distribution data may be traffic signs or road markings. We may like for the vehicle to exhibit cautious behaviour when anomalous input is detected, e.g. reducing speed when there is fog over the road (an OOD input). By perpetrating an attack on the vehicle's input feed, an adversary may be able to change the OOD input to be perceived as ID. This would cause the detection mechanism to fail to flag the input as anomalous, and the appropriate caution will not be taken. More real-world attack scenarios are given in Appendix \ref{apdx:real_world}.

Owing to the importance of this problem, there have been a number of recent methods \cite{chen2021atom, chen2022robust,azizmalayeri2022your,shao2022open,hein2019relu} aiming to produce an \textit{adversarially robust OOD detection method}. These methods have demonstrated encouraging results, showing reduced susceptibility to attacks on their detection performance.

Despite these advances, current methods often struggle to: (a) maintain competitive \textit{clean} OOD detection performance (in the absence of an attacker), (b) be robust to strong attacks targeting \textbf{both} ID and OOD data, and (c) achieve strong ID classification accuracy. Empirically, many techniques that improve robustness to one kind of attack do not exhibit robustness to other kinds of attack \cite{chen2021atom}, and for methods that do improve robustness to multiple attacks, there is often a trade-off with other desirable properties \cite{zhang2019theoretically,chen2022robust, shao2022open}. Such limitations reduce the practicality and usefulness of these existing methods.

Given the substantial room for improvement in OOD detection, particularly under adversarial conditions, we aim to better understand the limitations of existing methods and identify key factors affecting performance. Through a motivating example, we demonstrate the necessity of training with adversarial attacks on both inlier and outlier data to achieve broad robustness. Furthermore, drawing on the established relationship between ID accuracy and OOD detection \cite{vaze2022openset}, we leverage insights from recent work aimed at improving the trade-off between clean and robust classification accuracy \cite{zhang2019theoretically,rade2022reducing} to enhance the analogous trade-off for clean and robust OOD detection.

In this paper, we propose a new technique for robust OOD detection: \textbf{HALO} (\textbf{H}elper-based \textbf{A}dversaria\textbf{L} \textbf{O}OD detection). We adapt the TRADES framework \cite{zhang2019theoretically} to the problem of OOD detection, discovering a novel objective function involving both inlier and outlier attacks. We additionally employ a helper model \cite{rade2022reducing} to internalise more appropriate decision boundaries, showing that this appreciably improves performance. Experimentally, we demonstrate that our method achieves state-of-the-art performance across a number of attack settings and datasets (Figure \ref{fig:radar}).  Concretely, compared to the next best method, over the datasets in the OpenOOD benchmark \cite{yang2022openood}, we achieve an average AUROC improvement of 3.15 on clean data, 8.99 under ID$\rightarrow$OOD attacks, 6.79 under OOD$\rightarrow$ID attacks and 5.44 under both attacks at once. Additionally, as our method has individual terms relating to each facet of performance, it has the advantage of being tuneable via hyperparameters to achieve the performance profile most suited to the given scenario.  

Our contributions are as follows: 
\begin{itemize}
    \item We introduce HALO, a simple method for instilling robustness to attacks on classification and detection.
    \item We show that our method exhibits strong, tuneable performance, outperforming existing methods and setting new SOTA performance across a number of datasets and attack settings.
    \item We show that HALO's improvement is robust, testing it against transferred attacks, against attackers with different strengths and in settings where extra data is available during training.
\end{itemize}

We hope that this work enables greater trust of OOD detection systems in real-world, safety critical settings.

\section{Background and Related Work}\label{sec:related}
In this section, we give an overview of the fields of OOD detection and adversarial machine learning, highlighting techniques relevant to our approach. We then review adversarial OOD detection and describe a number of prior robust OOD detection methods.

\subsection{Out-of-Distribution Detection}

The task of out-of-distribution (OOD) detection involves distinguishing between inputs drawn from the training distribution (in-distribution or ID samples) and those drawn from other distributions (out-of-distribution or OOD samples) \cite{nguyen2015deep}. Many approaches have been proposed to perform this task, including score-based approaches where a higher score is more indicative of an OOD sample. Broadly, score-based approaches can be broken down into confidence based \cite{hendrycks17baseline,liu2023gen,liang2018enhancing,wang2022vim}, energy based \cite{liu2020energy,zhang2022out} and deriving from a learned discriminator \cite{azizmalayeri2022your,kong2021opengan}. Other methods focus on measuring distance between samples in normal \cite{lee2018simple,ren2021simple} and hyper-spherical embedding spaces \cite{ming2023how}.
Another line of work has explored giving models access to outlier samples from an 
auxiliary training set $\mathcal{D}_{\text{oe}}$ \cite{hendrycks2018deep}. This method, called outlier exposure, works by training on the following objective:
\begin{equation}
    \mathcal{L}_{\text{OE}} = \mathbb{E}_{(\bm{x},y) \sim \mathcal{D}_{\text{in}}} [\text{CE}(f_\theta(\bm{x}), y)] + \lambda \cdot \mathbb{E}_{\bm{x}' \sim \mathcal{D}_{\text{oe}}} [\text{KL} (f_\theta(\bm{x}') , \mathcal{U})]
    \label{eq:outlier_exposure}
\end{equation}
where the standard cross-entropy (CE) loss term is paired with a KL divergence term that pushes output on OE samples towards the uniform distribution. Recent work has also explored generating synthetic outliers \cite{duvos,taonon} and using large pre-trained image-text models \cite{wang2023clipn} to improve performance.

\subsection{Adversarial Attacks and Defences} \label{sec:advml}
Adversarial attacks are small, often imperceptible modifications to inputs to a model that change the class label that the model assigns to the input  \cite{Szegedy2014}. The field of adversarial machine learning is concerned both with developing new methods of attack as well as techniques to produce robust models that perform better under these attacks.

\textbf{Adversarial Attacks:} An adversarial example for an image $\bm{x}$ with class label $y$ is defined as $\bm{x}' = \bm{x} + \bm{\delta}$, where $\bm{\delta}$ is a perturbation that satisfies some constraint designed to ensure its imperceptibility to humans, usually that the $\ell_p$ norm of delta $||\bm{\delta}||_p$ is less than some $\epsilon$. We denote the set of such allowable perturbations $\mathcal{S}$. Perturbations are usually constructed to maximise the loss function $\text{CE}(f_\theta(\bm{x}+\bm{\delta}), y)$, thus lowering classification accuracy. 

Projected gradient descent (PGD) \cite{madry2018towards} is a common method used to produce adversarial perturbations. It constructs $\bm{\delta}$ by making a number of additive gradient steps of size $\alpha$ in the direction of maximum loss:
\begin{equation}
\bm{\delta}^{(t+1)}_{\text{PGD}} = \Pi_{\mathcal{S}} \bigl( \bm{\delta}^{(t)}_\text{PGD} + \alpha \cdot \text{sign}(\nabla_{\bm{x}} \text{CE}(f_\theta(\bm{x} + \bm{\delta}^{(t)}_{\text{PGD}}), y) \bigr)
\label{eq:pgd_delta}
\end{equation}
where $\Pi_{\mathcal{S}}$ denotes projection onto the set of allowed perturbations $\mathcal{S}$.

AutoAttack \cite{croce2020reliable} is a more recent, stronger method that comprises four independent attacks and is widely considered to provide a more reliable estimate of robustness against various types of adversaries.

\textbf{Adversarial Defences:} In parallel to the development of adversarial attacks, many works have proposed methods to produce \textit{adversarially robust} classifiers \cite{goodfellow2015explaining,madry2018towards,zhang2019theoretically,rade2022reducing}. This task has proven difficult, with all methods that are effective in increasing robust accuracy resulting in significant reductions in clean accuracy \cite{tsipras2018robustness,raghunathan2020understanding}. As such, a primary focus of research in this area has been to optimise the trade-off between robust accuracy and clean accuracy. An early work that achieves robustness to various attacks is standard adversarial training (SAT) \cite{madry2018towards}. This method treats training as a min-max game, optimising the normal cross-entropy objective in the presence of a simulated adversary. Specifically, AT aims to minimise: 
\begin{equation}
\mathcal{L}_{\text{AT}} = 
\mathbb{E}_{(\bm{x},y)\sim\mathcal{D}} \bigl[ \max_{\bm{\delta} \in \mathcal{S}} \text{CE}(f_\theta(\bm{x} + \bm{\delta}), y) \bigr]
\label{eq:madry_at}
\end{equation}
where the inner maximisation is approximated using a PGD attack. \citet{zhang2019theoretically} leverage a better theoretical understanding of the trade-off between clean and robust accuracy to develop TRADES, an improvement on SAT. TRADES has separate terms for clean accuracy and robustness, using a hyperparameter $\beta$ to control their relative importance:
\begin{equation}
\begin{split}
\mathcal{L}_{\text{TRADES}} &= 
\mathbb{E}_{(\bm{x},y)\sim\mathcal{D}} \bigl[ \text{CE}(f_\theta(\bm{x}), y) \\
&+ \beta \cdot \max_{\bm{\delta} \in \mathcal{S}} \text{KL} (f_\theta(\bm{x}) , f_\theta(\bm{x} + \bm{\delta})) \bigr]
\label{eq:trades}
\end{split}
\end{equation}

Helper Adversarial Training (HAT) \cite{rade2022reducing} is a more recent method that attempts to further improve the clean/robust trade-off by reducing the excessive decision boundary margin that is introduced by existing adversarial training methods. It works by defining a helper example $\Tilde{\bm{x}} = \bm{x} + 2 \bm{\delta}$ and helper label $\Tilde{y} = \argmax f_{std}(\bm{x}+\bm{\delta})$ where $f_{std}$ is a standard (non-robust) model. We can then optimise the following objective: 
\begin{equation}
\begin{split}
\mathcal{L}_{\text{HAT}} &= 
\mathbb{E}_{(\bm{x},y)\sim\mathcal{D}} \bigl[ \text{CE}(f_\theta(\bm{x}), y) \\
&+ \beta \cdot \max_{\bm{\delta} \in \mathcal{S}} \text{KL} (f_\theta(\bm{x}) , f_\theta(\bm{x} + \bm{\delta})) \\
&+ \gamma \cdot \text{CE}(f_\theta(\Tilde{\bm{x}}), \Tilde{y}) \bigr]
\label{eq:hat}
\end{split}
\end{equation}

This has the effect of encouraging the model to be invariant in its output to attacks of the specified strength $\epsilon$, but less so beyond that. HAT is shown to improve the trade off between clean and robust accuracy compared to TRADES, while achieving a similar robust accuracy \cite{rade2022reducing}. Recently, other works have explored mixing standard and robust model output \cite{bai2024improving} and incorporating extra data from pseudo-labelling \cite{carmon2019unlabeled,gowal2020uncovering} or diffusion models \cite{wang2023better} to achieve better results.

\subsection{Adversarial OOD Detection} \label{sec:advood}
Though not as well studied as the problems of (clean) OOD detection and (classification) adversarial robustness, the field of adversarial OOD detection has received meaningful attention in recent years \cite{azizmalayeri2022your,chen2021atom,shao2022open}.

Unlike \textit{classification} attacks in which the adversary aims to induce misclassification, detection attacks instead try to fool the detection module of a network. These attacks can naturally be separated into two types: \textbf{ID$\rightarrow$OOD} attacks which modify ID samples in an attempt to have the detector erroneously flag them as OOD, and \textbf{OOD$\rightarrow$ID} attacks which modify OOD samples to evade detection by the model. For clarity, we give a schematic diagram of each of these attacks in Figure \ref{fig:attack_vis}. We can also consider performing OOD detection when both of these attacks occur at the same time, which as noted by \citet{azizmalayeri2022your} is a strictly harder setting as the perturbation budget is effectively doubled. As approaches to OOD detection vary, the exact method used by the adversary also needs to change to produce a strong attack. When using the baseline method of MSP for example, it is desirable for the model's output to have a high entropy (uniform) distribution on OOD samples and a low entropy (peaked) distribution on ID samples. Correspondingly, OOD$\rightarrow$ID and ID$\rightarrow$OOD attacks for this method aim to increase/decrease the divergence between the model's output distribution and the uniform distribution, respectively. Details for other methods are given in Appendix \ref{apdx:detection_attacks}.

\subsection{Prior Robust OOD Detection Works}

ACET \cite{hein2019relu} is one of the first methods to tackle the problem of robust OOD detection, training jointly on standard CE loss and an (attacked) OE objective: 
\begin{equation}
\begin{split}
    \mathcal{L}_{\text{ACET}} &= 
    \mathbb{E}_{(\bm{x},y) \sim \mathcal{D}_{\text{in}}} \bigl[\text{CE}(f_\theta(\bm{x}), y)\bigr] \\
    &+ \lambda \cdot \mathbb{E}_{\bm{x}' \sim \mathcal{D}_{\text{oe}}} \bigl[\max_{\bm{\delta} \in \mathcal{S}}\text{CE}(f_\theta(\bm{x}'+\bm{\delta}), \mathcal{U})\bigr]
    \label{eq:ACET}
\end{split}
\end{equation}
where an OOD$\rightarrow$ID attack is performed using PGD to find $\bm{\delta}$. ATOM \cite{chen2021atom} takes a somewhat similar approach, training a classification model with $K+1$ classes, where a positive detection is made when $\textsc{SoftMax}(f_\theta(\bm{x}))_{K+1} \geq \gamma$ for some threshold $\gamma$. Training utilises a supplementary OE dataset where a subset of the samples are attacked to push them closer to the detection decision boundary. Both ACET and ATOM only consider robustness on OOD samples, leaving them vulnerable to ID$\rightarrow$OOD attacks. ALOE \cite{chen2022robust} and RATIO\footnote{Both ALOE and RATIO optimise the same objective function, but consider the case of the $\ell_\infty$ and $\ell_2$ norm respectively. In our comparisons we use ALOE as we also focus on the $\ell_\infty$ norm.} \cite{augustin2020adversarial} seek to rectify this issue, attacking both the ID and OE datasets during training. This amounts to optimising the following objective, where $\lambda$ is a hyperparameter controlling the relative importance of the OE term: 
\begin{equation}
\begin{split}
    \mathcal{L}_{\text{ALOE}} &= 
    \mathbb{E}_{(\bm{x},y) \sim \mathcal{D}_{\text{in}}} \bigl[\max_{\bm{\delta} \in \mathcal{S}}\text{CE}(f_\theta(\bm{x}+\bm{\delta}), y)\bigr] \\
    &+ \lambda \cdot \mathbb{E}_{\bm{x}' \sim \mathcal{D}_{\text{oe}}} \bigl[\max_{\bm{\delta} \in \mathcal{S}}\text{CE}(f_\theta(\bm{x}'+\bm{\delta}), \mathcal{U})\bigr]
    \label{eq:aloe}
\end{split}
\end{equation}

Another method aiming at both ID and OOD robustness is OSAD \cite{shao2022open} which trains an encoder with dual-attentive feature-denoising layers and uses the representations it produces as input to a classification head. This is paired with an OpenMax \cite{bendale2016openmax} layer for OOD detection. Further, Adversarially Trained Discriminator (ATD) \cite{azizmalayeri2022your} is a recent method that utilises a pre-trained robust classification model for feature extraction and trains a discriminator to perform OOD detection on these features. It additionally uses an OE dataset and a generator (trained in tandem with the discriminator) to gain exposure to a diverse distribution of outliers during training. 

Other recent works have explored certified OOD detection \cite{meinke2022provably}, robust one class classification \cite{bethune2023robust} and the use of pre-trained image-text models to supplement OE data \cite{mirzaeirodeo}.

\section{Proposed Method}
\label{halo}

This section formally introduces our method, called \textbf{H}elper-based \textbf{A}dversaria\textbf{L} \textbf{O}OD Detection (\textbf{HALO}). We first give a formal problem statement, introducing the threat model and notation used throughout the paper. We then introduce the formulation that constitutes HALO, describing the motivation for each component part. Finally, we discuss the benefits of HALO over other methods.

\subsection{Problem Statement}

Let $\mathcal{D}_{\text{in}}$, $\mathcal{D}_{\text{out}}$, and $\mathcal{D}_{\text{oe}}$ be the in-distribution (ID), out-of-distribution (OOD), and outlier exposure (OE) distributions respectively, all defined over an input space $\mathcal{X} \subset \mathbb{R}^d$. We denote by $\mathbf{X} \in \mathbb{R}^d$ a random input vector with realisation $\bm{x}$, $Y \in \{-1,1\}$ a binary class label, $Y_{det} \in \{-1,1\}$ an OOD detection label (1 for OOD, -1 for ID), and $\mathbf{Y} \in \{0,1\}^K$ a one-hot encoded label for $K$-class problems with realisation $\bm{y}$.

Let $f_\theta: \mathbb{R}^d \rightarrow \mathbb{R}^K$ denote a neural network classifier parameterised by $\theta$ that takes an input $\bm{x}$ and outputs a softmax distribution $\bm{z} = f_\theta(\bm{x})$ where $\sum_i z_i = 1$. For OOD detection, we consider a detector $g: \mathbb{R}^d \rightarrow \mathbb{R}$ that assigns a scalar OOD score to each input. An input $\bm{x}$ is classified as OOD if $g(\bm{x}) > \tau$ for some threshold $\tau$. 

Additionally, we use $\mathcal{U}$ to denote the uniform distribution over $K$ classes, $\text{KL}$ to denote the Kullback-Leibler divergence and $\text{CE}$ to denote the cross entropy loss. 

In the context of adversarial attacks on OOD detection, we consider two types of attacks:

\begin{enumerate}
    \item \textbf{ID$\rightarrow$OOD attacks}: For $\bm{x} \in \mathcal{D}_\text{in}$, find $\bm{\delta}$ such that $\|\bm{\delta}\|_p \leq \varepsilon$ and $g(\bm{x} + \bm{\delta}) > \tau$,
    \item \textbf{OOD$\rightarrow$ID attacks}: For $\bm{x} \in \mathcal{D}_\text{out}$, find $\bm{\delta}$ such that $\|\bm{\delta}\|_p \leq \varepsilon$ and $g(\bm{x} + \bm{\delta}) \leq \tau$,
\end{enumerate}
where $\|\cdot\|_p$ denotes the $\ell_p$ norm and $\varepsilon$ is the maximum allowed perturbation magnitude. 

We use $\mathbf{B}(\bm{x},\epsilon)$ to denote the set $\{\bm{x}' \in \mathcal{X}: \| \bm{x}-\bm{x}' \|_p < \epsilon \}$, i.e., the points around a sample $\bm{x}$ within the $\epsilon$-ball. We use $\mathcal{S}$ to denote the special case where $\bm{x} = \mathbf{0}$, i.e., the set of allowable perturbations.

We denote the classifier's decision boundary $DB(f) = \{\bm{x} \in \mathcal{X}: f(\bm{x})=\mathbf{0} \}$. 
We further denote the neighbourhood around $DB(f)$ reachable by an allowable perturbation $\mathbf{B}(DB(f),\epsilon) = \{\bm{x} \in \mathcal{X}: \exists \bm{x}' \in \mathbf{B}(\bm{x},\epsilon) \ \text{s.t.}\  f(\bm{x})f(\bm{x}') \leq 0 \}$.

Our goal is to find $f$ and $g$ such that $g$ is robust to both types of detection attack while maintaining high clean OOD detection performance and that $f$ has high classification accuracy.

\subsection{Theoretical Framework}\label{sec:method}

In order to simultaneously achieve all desiderata above, we jointly optimise terms for classification \textit{and} detection. Our key insights in this section are:
\begin{enumerate}[label=\textbf{\alph*})]
    \item We can adapt an existing adversarial training framework, TRADES \cite{zhang2019theoretically}, to the task of OOD detection.
    \item By making a principled choice of detector $g$ we can recover an optimisation problem in terms of $f$ only. 
    \item Introducing an additional helper term to the optimisation problem comprehensively improves performance (as motivated in Section \ref{sec:helper_explanation}).
\end{enumerate}

\textbf{Framework Adaptation:} A commonly used strong adversarial training algorithm is TRADES \cite{zhang2019theoretically}. In this work, the authors decompose the robust error of a binary classifier $f$ into the sum of its natural (clean) error and its boundary error (points that could be successfully attacked): $\mathcal{R}_{rob}(f) = \mathcal{R}_{nat}(f) + \mathcal{R}_{bdy}(f)$. They show that the difference between robust error and optimal clean error is upper bounded by the probability that a point in $\mathcal{D}_{\text{in}}$ lies within the set $\mathbf{B}(DB(f),\epsilon)$ and is correctly classified by $f$ \cite[Theorem 3.1]{zhang2019theoretically}. This suggests that to minimise the difference between robust and natural accuracy, we should minimise:
$$\phi(f(\mathbf{X})Y) + \max_{\bm{\delta} \in \mathcal{S}}\phi(f(\mathbf{X})f(\mathbf{X}+\bm{\delta})/\lambda)$$
where $\phi$ is a classification calibrated loss \cite{bartlett2006convexity}. Considering detection attacks rather than classification attacks, we can \textbf{adapt the TRADES framework} by making some substitutions. We replace the classifier $f$ with a detector $g'$, constructed by $g'(x) = g(x) -\tau$ such that $g'(X) > 0$ represents a positive detection and $g'(X) < 0$ represents a negative detection. We also use $Y_{det}$ in place of $Y$. Then, we can follow the same chain of reasoning, finding a similar upper bound in terms of $\mathbf{B}(DB(g'),\epsilon)$ which suggests minimising: 
$$\phi(g'(\mathbf{X})Y_{det}) + \max_{\bm{\delta} \in \mathcal{S}}\phi(g'(\mathbf{X})g'(\mathbf{X}+\bm{\delta})/\lambda)$$
\textbf{Choosing $\bm{g}$:} The choices of $g(\bm{x})$ and $\phi$ are free parameters in this optimisation problem, however, we choose $g(\bm{x}) = H(f_\theta(\bm{x}))$ (the Shannon entropy of the softmax distribution\footnote{We make this choice following the work of \citet{hendrycks2018deep}, where this optimisation target is used.}) and $\phi(\alpha) = exp(-\alpha)$ (the exponential loss). These choices are made so that we can use an entropy-based post-processor for OOD detection. Expressing $g$ in terms of $f$ like this allows us to optimise the weights of a \textbf{single model}. As $Y_{det} = 1$ in the case of OE input, we can see that $\phi(g'(\mathbf{X})Y_{det}) = exp(\tau-H(f_\theta(\mathbf{X})))$, and so minimising this term is equivalent to maximising $H(f_\theta(\mathbf{X}))$ which is in turn equivalent to minimising $\text{KL}(f_\theta(\mathbf{X}),\mathcal{U})$ (See Appendix \ref{apdx:proofs} for a full proof). For the regularisation term, we would like to encourage $g'(\mathbf{X})$ and $g'(\mathbf{X}+\bm{\delta})$ to be similar, which for our chosen $g$, means encouraging the entropy of $f_\theta(\mathbf{X})$ and $f_\theta(\mathbf{X}+\bm{\delta})$ to be similar. Instead of optimising directly for similar entropy, we find that the heuristic approach adopted by \citet{zhang2019theoretically} of minimising: $$\text{KL}(f_\theta(\mathbf{X}),f_\theta(\mathbf{X}+\bm{\delta}))$$ also works well for our problem. Note that minimising this term is not exactly equivalent to minimising differences in entropy, instead, it minimises differences across the whole distribution. In practice, we find that it does a good job of introducing entropy invariance and has the advantage of being the same as the ID robustness term, which simplifies implementation.

\subsection{Joint Optimisation}

In order to achieve robust performance on both inlier and outlier data, we jointly optimise objectives for both classification and detection. As per our choices above, these objectives are \textbf{both} in terms of $f(\bm{x})$. To extend to the domain of multi-class classification, we follow \citet{zhang2019theoretically} and use the cross entropy loss, as it is also classification calibrated \cite{bartlett2006convexity}. We note that the objective function for $g$ is only optimised on OE data ($Y_{det} = 1$) as the optimisation of $f$ on ID data already leads to low entropy on the ID distribution through the cross-entropy term. Putting these two components together, we optimise the sum of the classification and detection losses on their respective datasets:
\begin{equation}
\begin{split}
    \mathcal{L}_{\text{HALO}} = \mathbb{E}_{(\bm{x},\bm{y})\sim\mathcal{D}_{\text{in}}} \bigl[\ell_{\text{ID}}(\bm{x},\bm{y})\bigr] + \eta \cdot \mathbb{E}_{\bm{x}' \sim \mathcal{D}_{\text{oe}}} \bigl[ \ell_{\text{OE}}(\bm{x}') \bigr] 
\end{split}
\label{eq:halo}
\end{equation}
where:
\begin{equation}
\begin{split}
    \ell_{\text{ID}} =  \text{CE}(f_\theta(\bm{x}), \bm{y}) &+ \beta_1 \cdot \max_{\bm{\delta} \in \mathcal{S}} \text{KL} (f_\theta(\bm{x}) , f_\theta(\bm{x} + \bm{\delta})) \\
    &+ \gamma \cdot \text{CE}(f_\theta(\Tilde{\bm{x}}), \Tilde{\bm{y}}) 
\end{split}
\label{eq:l_id}
\end{equation}

\begin{equation}
\begin{split}
    \ell_{\text{OE}} = \text{KL}(f_\theta(\bm{x}), \mathcal{U}) + \beta_2 \cdot \max_{\bm{\delta} \in \mathcal{S}} \text{KL} (f_\theta(\bm{x}) , f_\theta(\bm{x} + \bm{\delta}))
\end{split}
\label{eq:l_oe}
\end{equation}

There are four hyperparameters, each of which control the relative importance of an objective term. $\eta$ corresponds to the importance of the OE loss term, $\gamma$ corresponds to the importance of the helper loss term, and $\beta_1$ and $\beta_2$ correspond to the importance of the ID and OOD robustness terms, respectively. For simplicity, we always set $\beta_1 = \beta_2 = \beta$ in our experiments unless otherwise specified (such that the coefficient of the OE robustness term in expanded form is $\eta\cdot\beta$). Note also that $\Tilde{\bm{x}}$ and $\Tilde{\bm{y}}$ have the same meaning as in the HAT objective function (see Section \ref{sec:advml} for details on these terms). To construct the OOD detector $g$, we can choose any entropy-based post-processor. In this work, we choose a recent method: GEN \cite{liu2023gen} by default as it has been shown to be similar to Shannon entropy but more sensitive to small deviations, resulting in increased discriminative power.

\subsection{Intuitive Explanation}
Here, we give an intuitive explanation of the first two terms in both $\ell_{\text{ID}}$ (Equation \ref{eq:l_id}) and $\ell_{\text{OE}}$ (Equation \ref{eq:l_oe}). The inclusion of the helper loss (third term in $\ell_{\text{ID}}$) is explained in the next section.

As in the original TRADES formulation, the first term of $\ell_{\text{ID}}$: $\text{CE}(f_\theta(\bm{x}), \bm{y})$ aims to increase the clean accuracy of the model, whereas the second term aims to improve robustness to adversaries by reducing the amount by which they can change the distribution. Looking at $\ell_{\text{OE}}$, minimising the first term: $\text{KL}(f_\theta(\bm{x}), \mathcal{U})$ corresponds to maximising entropy on the OE distribution. Increasing entropy on the OE distribution, $\mathcal{D}_{\text{oe}}$, is done in the hopes of causing model output on the test-time OOD distribution, $\mathcal{D}_{\text{out}}$, to have higher entropy, allowing the detector $g$ to better distinguish between sample types. The second term: $\text{KL} (f_\theta(\bm{x}) , f_\theta(\bm{x} + \bm{\delta}))$ is the same as the second term in $\ell_{\text{ID}}$ and is concerned with improving robustness to perturbed data. As noted, we do not directly minimise the difference in entropy between these distributions, instead we minimise the KL divergence, which we find to be a good proxy. Heuristically, we note that as the KL divergence between these distributions approaches 0, they tend to become more similar and hence similar in entropy. In the limit where the KL divergence = 0, the distributions are the same and therefore have the same entropy. 

We also note that even though the same attack targeting the KL divergence term is used in both loss functions $\ell_{\text{ID}}$ and $\ell_{\text{OE}}$, the differing first terms means that the resultant attack is qualitatively different. In the case of the ID loss term, the model's output on the natural input is pushed to be closer to the one-hot vector $\bm{y}$, whereas for the OE loss term it is pushed to be closer to the uniform distribution $\mathcal{U}$. This means that as the model converges, the training adversary will effectively be performing ID$\rightarrow$OOD and OOD$\rightarrow$ID attacks, respectively, providing a strong and varied signal to learn from. In Section \ref{sec:toy} we explore a toy model that verifies the importance of including both of these robustness terms, showing that optimising only one does not produce meaningful robustness to the other kind of attack. We also note that optimising these two terms can be thought of as performing semi-supervised training on the other dataset, as alluded to in Section 4 of \citet{zhang2019theoretically}

\subsection{Inclusion of Helper Loss} \label{sec:helper_explanation}

To explain the inclusion of the helper loss term, $\text{CE}(f_\theta(\Tilde{\bm{x}}), \Tilde{\bm{y}})$, we review the relationships between clean accuracy, robust accuracy and detection performance. 
\citet{vaze2022openset} study the relationship between classification accuracy and performance in open-set recognition, a task closely related to OOD detection. They show that detection metrics such as AUROC and FPR95 are highly correlated with classification accuracy, and suggest that: `a good closed-set classifier is all you need'. Concurrently, it has been well documented that adversarial training significantly degrades clean accuracy \cite{tsipras2018robustness,raghunathan2020understanding}. Taken together, these two findings imply that in making our method robust to adversarial attacks we may end up sacrificing both clean ID accuracy, and clean OOD detection performance. They also imply that if we can improve the trade-off between clean and robust ID accuracy, we may get better detection performance `for free'. To this end, we include the helper loss term introduced by \citet{rade2022reducing} in our ID loss function, which is shown to improve the clean/robust trade-off (see Section \ref{sec:advml} for a review of this method).

\section{Classification Robustness is not Sufficient for Robust OOD Detection} \label{sec:toy}

A natural question to ask is whether established adversarial training techniques such as SAT or TRADES are sufficient to defend against detection attacks. In this section we present a motivating example that suggests that:
\begin{enumerate}[label=\alph*)]
\item Classification AT methods are insufficient for robust OOD detection, and
\item To develop appropriate OOD robustness, a model must be exposed to attacks on the ID \textbf{and} OE datasets during training. 
\end{enumerate}

\begin{figure*}[t!]
\centering
\includegraphics[width=0.95\textwidth]{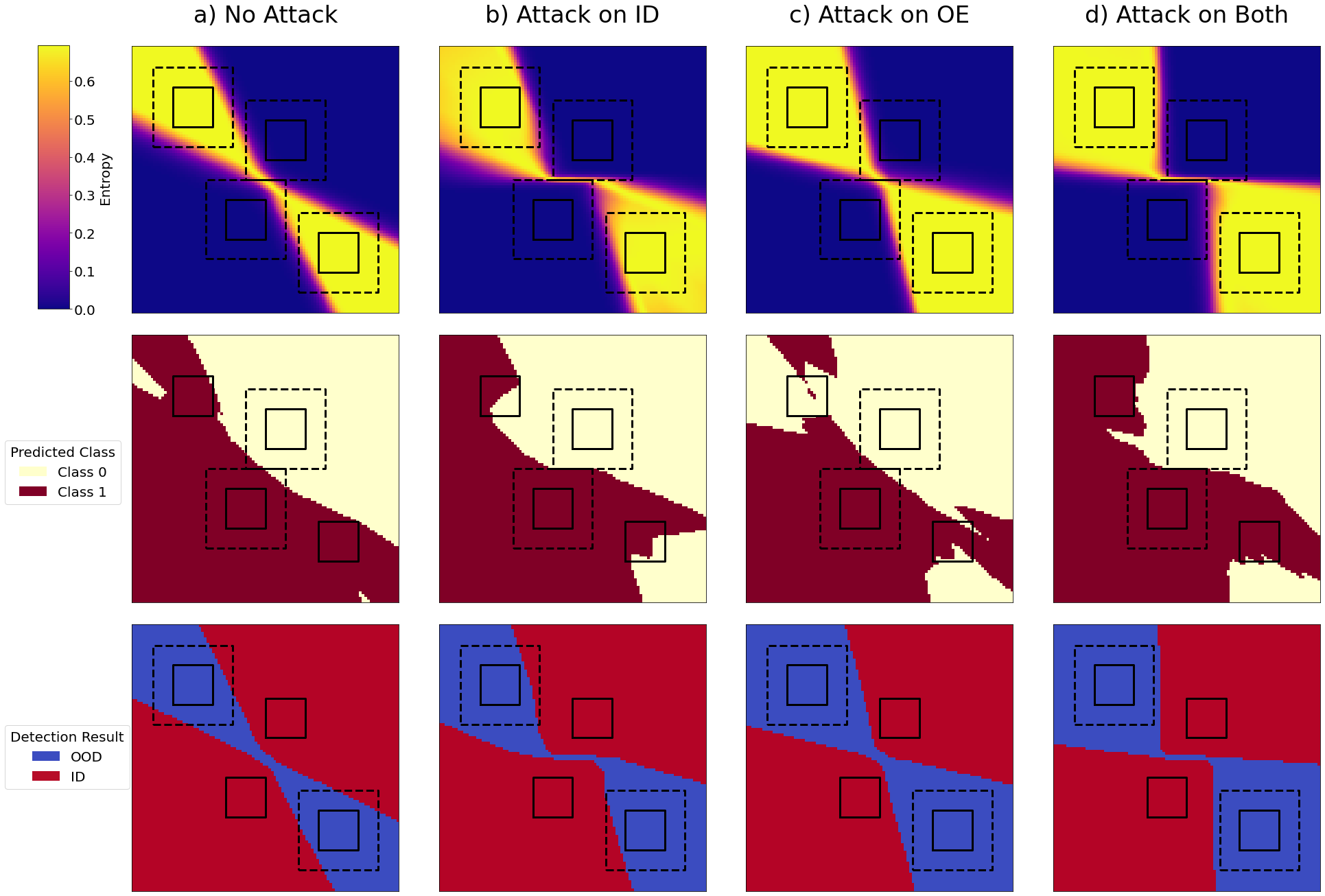}
\caption{Two-dimensional toy model showing the effect of different adversarial training methods. Solid lines represent training data distribution, dashed lines represent allowable perturbations. Training methods left to right: (a) Standard training with OE, (b) Adversarial training on ID data, (c) Adversarial training on OE data, (d) Adversarial training on both ID and OE data. Figures top to bottom: entropy distribution of possible inputs, classification decision boundary, detection decision boundary. Only model (d) that trains on both types of attack develops robust classification \textbf{and} detection decision boundaries.}
\label{fig:toy}
\end{figure*}

\subsection{Toy Model Setup} 
We demonstrate the effects of different adversarial training approaches using a simple binary classification task in a 2D space, as shown in Figure \ref{fig:toy}. There are four regions of interest in the space, represented by solid black squares. The two center regions represent two ID class distributions (class 0 and class 1) and the two regions in the upper left and lower right represent OE data. Dashed lines around these regions represent the set of reachable perturbed inputs (within the $\epsilon$-sized $\ell^{\infty}$ ball). Importantly, we note that it is possible to learn a model that achieves 100\% robust classification and detection accuracy for these data distributions. 

For our classifier, we train a simple MLP model with two hidden layers of size 64 using SGD with a learning rate of 0.001. We train four different models:
\begin{enumerate}[label=\alph*)]
\item A standard model using the OE training objective (Equation \ref{eq:outlier_exposure}).
\item The standard model with an additional TRADES robustness term for ID data as in Equation \ref{eq:trades}. 
\item The standard model with the same robustness term, but for OE data only.
\item A model with both ID and OE robustness terms.
\end{enumerate}

For all of these models, training data consists of 1000 points randomly sampled from within the Class 0, Class 1 and OE regions. See Appendix \ref{apdx:toy_model} for the full toy model details. 

\subsection{Results}
As shown in Figure \ref{fig:toy}, for each model we plot: the entropy of the softmax distribution, the classification decision boundary, and the detection decision boundary (constructed using MSP with threshold $\tau=0.9$). We can see in sub-figure a) that standard training produces a classifier that has 100\% clean accuracy and perfect detection, however it is clear that adversarial attacks are able to push certain data-points over both the classification and detection decision boundaries. In sub-figures b) and c) we see the results of additionally performing adversarial training on either the ID or OE sets respectively. In these cases the model has developed robustness to the type of attack it was trained in the presence of, but not to the other attack type. Taking model b) as an example, we can see that though the entire dashed regions around the ID distributions are being assigned the correct class, the regions around the OE distributions contain samples that evade detection as OOD. In sub-figure d) we can see the result of training with attacks on both data distributions. The model exhibits perfect robust classification \textit{and} detection boundaries, correctly classifying/detecting all allowable perturbed inputs. 

This motivating example shows the failure of existing classification AT methods to perform robust OOD detection, suggesting that these two types of robustness are not interchangeable. We also note that attacking the OE data creates a larger region of uncertainty around the OE distribution, potentially improving generalisation to test time OOD samples. 

These results suggest that strong performance can be achieved by jointly training on attacked ID and OE data, as in our method in Section \ref{sec:method}. 

\section{Experiments}

In this section we run multiple experiments on HALO and other robust OOD detection methods to better understand the nature of their performance:
\begin{enumerate}
    \item We first test HALO and six other robust OOD detection methods in a benchmark setting with CIFAR-10 as the ID dataset. We show that HALO outperforms existing methods in almost all attacked settings, retaining strong clean performance (Section \ref{sec:comparison}).
    \item We study the sensitivity of our proposed method to different choices of hyperparameter settings, analysing both classification and OOD detection performance. We find that HALO's performance can be tuned via these hyperparameters, and that distinct optimal regions exist for different types of robustness (Section \ref{sec:sens}).
    \item We test HALO and the other methods against attacks of differing strengths, showing that HALO outperforms others regardless of attack strength. (Section \ref{sec:attack_params}).
    \item We test HALO's susceptibility to two black box attacks to ensure that its robustness is not merely due to a low strength attack. (Section \ref{sec:tfr}, results in Appendix \ref{apdx:blackbox_details}).
    \item We compare our method to another strong method in the setting where extra pseudo-labelled data is available, showing that it retains its advantage (Section \ref{sec:extra_data}).
    \item We explore different choices of post-processor that can be paired with HALO and show their impact on detection performance in clean and robust settings, showing that GEN \cite{liu2023gen} is the strongest performer (Section \ref{sec:ppstudy}).
\end{enumerate}

We direct the reader to Appendix \ref{apdx:ex_details} for the detailed experimental setups and to Appendix \ref{apdx:extra_results} for additional results and an ablation study.

\subsection{Details of Experimental Settings} \label{sec:eval}

In this section, we provide the details of our experimental
settings.

\subsubsection{Training Setup} To ensure fairness and comparability, we use a consistent training setup for each method. We train a ResNet-18 \cite{he2016deep} following the training procedure described in \citet{rade2022reducing}: training for 50 epochs using SGD with Nesterov momentum \cite{nesterov1983method} and a batch size of 128 for both ID and OE datasets. Learning rates follow a cyclic schedule \cite{smith2019super} with cosine annealing and a maximum learning rate of 0.21. When performing adversarial training we again follow \citet{rade2022reducing}, using a PGD adversary with a maximum $\ell_\infty$ perturbation of $\epsilon=8/255$ and 10 steps of size $\alpha = 2/255$. We use adversarial validation \cite{rice2020overfitting} with the same PGD-10 adversary to select the final model used in evaluation. For ATD and HALO, we use the same training method to train the feature extractor and standard models, respectively. Unless otherwise stated, we set $\eta=2.0$, $\gamma=0.5$ and $\beta=3.0$. More details on the training setup are available in Appendix \ref{apdx:trainingsetup}.
    
\begin{table*}[t!]
\centering
\caption{Comparison of AUROCs across different datasets and attack settings for each method. Methods lacking both kinds of robustness are \textbf{\textcolor{gray}{greyed out}}. Best and second-best results among robust methods are \textbf{bolded} and \underline{underlined}, respectively. The $\Delta$ column shows the difference between HALO and the best other method, colored \textcolor{green}{green} if positive and \textcolor{red}{red} if negative. CIFAR-10 \cite{krizhevsky2009learning} is used as the ID dataset and all attacks are performed using PGD-40 with $\epsilon=8/255$. Methods that use a post-processor are paired with GEN \cite{liu2023gen}. Full details are available in Appendix \ref{apdx:trainingsetup}. On all datasets but MNIST, HALO has a) the highest AUROC in all adversarial settings and b) the highest clean AUROC of all methods robust to OOD and ID attacks.}
\begin{tabular}{llrrrrrrr|r}   
\hline
Dataset & Attack Type & \textcolor{gray}{OE} & \textcolor{gray}{ATOM} & TRADES & ALOE & OSAD & ATD & HALO & $\Delta$ \\
\hline
\multirow{4}{*}{CIFAR100} & Clean & \textcolor{gray}{90.63} & \textcolor{gray}{91.39} & 71.74 & 73.78 & 74.54 & \underline{80.11} & \textbf{83.44} & \textcolor{green}{+3.33} \\
& ID$\rightarrow$OOD & \textcolor{gray}{0.37} & \textcolor{gray}{3.33} & \underline{57.33} & 50.80 & 51.77 & 53.64 & \textbf{66.86} & \textcolor{green}{+9.53} \\
& OOD$\rightarrow$ID & \textcolor{gray}{0.18} & \textcolor{gray}{49.76} & 58.60 & 53.70 & 54.86 & \underline{60.49} & \textbf{67.51} & \textcolor{green}{+7.02} \\
& Both & \textcolor{gray}{0.01} & \textcolor{gray}{0.11} & \underline{44.45} & 31.93 & 33.23 & 34.95 & \textbf{48.85} & \textcolor{green}{+4.40} \\
\hline
\multirow{4}{*}{TIN} & Clean & \textcolor{gray}{99.97} & \textcolor{gray}{97.31} & 76.95 & 78.45 & 80.63 & \underline{86.31} & \textbf{90.68} & \textcolor{green}{+4.37} \\
& ID$\rightarrow$OOD & \textcolor{gray}{7.34} & \textcolor{gray}{6.30} & \underline{62.52} & 55.71 & 58.25 & 60.50 & \textbf{76.27} & \textcolor{green}{+13.75} \\
& OOD$\rightarrow$ID & \textcolor{gray}{0.46} & \textcolor{gray}{61.50} & 64.14 & 58.78 & 60.55 & \underline{68.47} & \textbf{76.70} & \textcolor{green}{+8.23} \\
& Both & \textcolor{gray}{0.01} & \textcolor{gray}{0.06} & \underline{49.21} & 35.76 & 37.66 & 40.64 & \textbf{57.39} & \textcolor{green}{+8.18} \\
\hline
\multirow{4}{*}{SVHN} & Clean & \textcolor{gray}{99.68} & \textcolor{gray}{98.89} & 80.11 & 79.84 & 86.49 & \underline{86.59} & \textbf{94.59} & \textcolor{green}{+8.00} \\
& ID$\rightarrow$OOD & \textcolor{gray}{4.24} & \textcolor{gray}{25.69} & \underline{64.72} & 54.88 & 63.56 & 58.63 & \textbf{80.94} & \textcolor{green}{+16.22} \\
& OOD$\rightarrow$ID & \textcolor{gray}{0.00} & \textcolor{gray}{51.34} & 67.30 & 59.07 & \underline{69.69} & 66.65 & \textbf{82.33} & \textcolor{green}{+12.64} \\
& Both & \textcolor{gray}{0.00} & \textcolor{gray}{0.03} & \underline{51.83} & 35.06 & 45.01 & 37.57 & \textbf{61.92} & \textcolor{green}{+10.09} \\
\hline
\multirow{4}{*}{Texture} & Clean & \textcolor{gray}{96.58} & \textcolor{gray}{98.84} & 73.73 & 76.55 & 82.32 & \underline{89.85} & \textbf{96.06} & \textcolor{green}{+6.21} \\
& ID$\rightarrow$OOD & \textcolor{gray}{2.46} & \textcolor{gray}{11.40} & 58.91 & 52.61 & 58.91 & \underline{67.41} & \textbf{85.40} & \textcolor{green}{+17.99} \\
& OOD$\rightarrow$ID & \textcolor{gray}{5.12} & \textcolor{gray}{76.81} & 63.00 & 58.88 & 67.31 & \underline{72.89} & \textbf{87.44} & \textcolor{green}{+14.55} \\
& Both & \textcolor{gray}{0.11} & \textcolor{gray}{0.58} & \underline{48.10} & 35.60 & 43.12 & 46.03 & \textbf{70.62} & \textcolor{green}{+22.52} \\
\hline
\multirow{4}{*}{Places365} & Clean & \textcolor{gray}{93.18} & \textcolor{gray}{96.77} & 78.08 & 79.59 & 82.27 & \underline{85.45} & \textbf{89.65} & \textcolor{green}{+4.20} \\
& ID$\rightarrow$OOD & \textcolor{gray}{0.52} & \textcolor{gray}{5.24} & \underline{64.10} & 57.91 & 60.86 & 59.86 & \textbf{74.85} & \textcolor{green}{+10.75} \\
& OOD$\rightarrow$ID & \textcolor{gray}{0.06} & \textcolor{gray}{59.18} & 63.95 & 58.55 & 60.38 & \underline{65.92} & \textbf{74.58} & \textcolor{green}{+8.66} \\
& Both & \textcolor{gray}{0.00} & \textcolor{gray}{0.04} & \underline{49.23} & 35.90 & 37.68 & 38.82 & \textbf{55.02} & \textcolor{green}{+5.79} \\
\hline
\multirow{4}{*}{MNIST} & Clean & \textcolor{gray}{99.35} & \textcolor{gray}{97.15} & 51.55 & 70.76 & 68.16 & \textbf{99.57} & \underline{92.33} & \textcolor{red}{-7.24} \\
& ID$\rightarrow$OOD & \textcolor{gray}{2.97} & \textcolor{gray}{4.85} & 38.31 & 46.57 & 49.68 & \textbf{96.63} & \underline{82.36} & \textcolor{red}{-14.27} \\
& OOD$\rightarrow$ID & \textcolor{gray}{0.01} & \textcolor{gray}{89.15} & 45.93 & 65.13 & 60.63 & \textbf{99.37} & \underline{89.02} & \textcolor{red}{-10.35} \\
& Both & \textcolor{gray}{0.00} & \textcolor{gray}{2.38} & 33.55 & 41.36 & 40.90 & \textbf{95.46} & \underline{77.10} & \textcolor{red}{-18.36} \\
\hline
\end{tabular}
\label{tab:main-results}
\end{table*}

\subsubsection{Evaluation Methodology and Datasets} Previous works \cite{hendrycks2018deep,augustin2020adversarial,azizmalayeri2022your} have presented results on a variety of different OOD and ID datasets and attack strengths, making comparison between methods difficult and potentially introducing bias. To address this issue, we adopt the dataset pairings outlined in the OpenOOD framework \cite{yang2022openood}. For most experiments we use CIFAR-10 as the ID dataset and TIN597 \cite[Appendix~A]{yang2022openood} as the OE dataset. We direct the reader to Appendix \ref{apdx:extra_results} for experiments with larger datasets. The six OOD datasets used with CIFAR-10 are: CIFAR-100 \cite{krizhevsky2009learning}, TinyImageNet \cite{le2015tiny} (disjoint with the OE set), MNIST \cite{MNIST}, SVHN \cite{netzer2011reading}, Textures \cite{cimpoi2014describing} and Places365 \cite{zhou2017places}. As well as ensuring fair evaluations, this has the advantage of allowing us to compare performance on near- and far-OOD tasks \cite{winkens2020contrastive}. 

To evaluate OOD detection methods, we primarily use  the area under the receiver operating characteristic curve (AUROC) \cite{davis2006relationship} but also give results using the false positive rate at a true positive rate of $95\%$ (FPR95) \cite{kumar2016learning} and area under the precision-recall curve (AUPR) \cite{manning1999foundations} in the appendix. 

When testing robust OOD detection performance, we use a PGD attack with $\epsilon=8/255$ and 40 steps of size $\alpha = 0.5/255$. For methods using standard post-processors (OE, TRADES, ALOE, OSAD and HALO) we use a PGD-style attack targeting increased/decreased entropy for ID$\rightarrow$OOD and OOD$\rightarrow$ID attack, respectively. For methods with unique detection methods, such as ATD and ATOM, we adapt our attack to properly target the given method. Full details of the various methods of attack are given in Appendix \ref{apdx:detection_attacks}. We use AutoAttack \cite{croce2020reliable} to test classification robustness, unless otherwise stated.

\subsection{Comparison of Different Methods} \label{sec:comparison}

\begin{figure*}[htb!]
\centering
\includegraphics[width=0.7\textwidth]{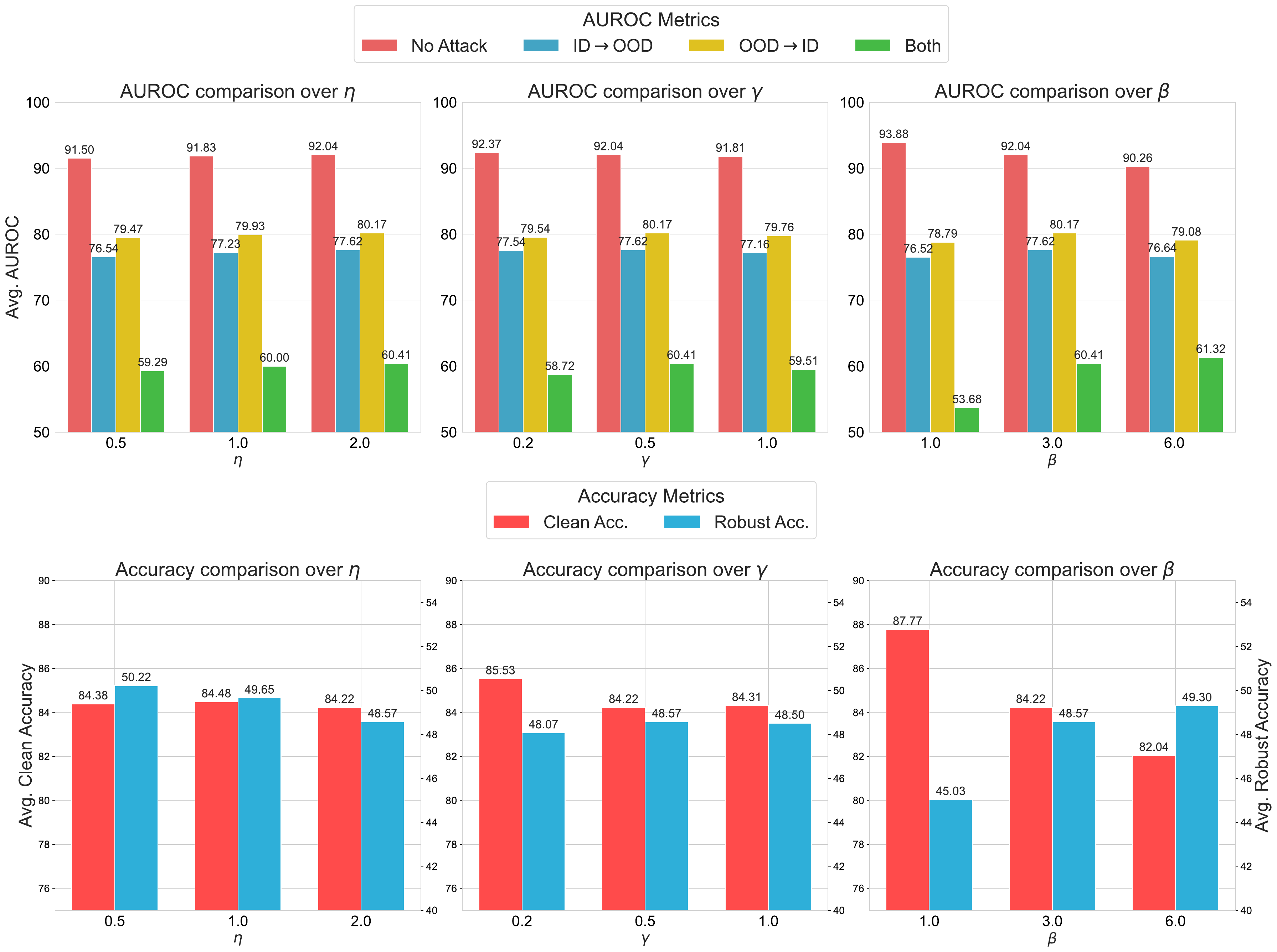}
\caption{Hyperparameter sensitivity analysis on CIFAR-10. By default we use $\eta=2.0$, $\gamma=0.5$ and $\beta = 3.0$ and only change the metric being examined. All results are averages over 3 independent runs. Top row: averaged AUROC across 6 datasets for different attack settings, bottom row: clean accuracy (left axis) and robust accuracy (right axis). Columns: varying $\eta$ (left), $\gamma$ (middle) and $\beta$ (right).}
\label{fig:sens}
\end{figure*}

To evaluate HALO, we compare it with six other methods: OE \cite{hendrycks17baseline}, ATOM \cite{chen2021atom}, TRADES \cite{zhang2019theoretically}, ALOE \cite{chen2022robust}, OSAD \cite{shao2022open} and ATD \cite{azizmalayeri2022your}. Table \ref{tab:main-results} describes the detection performance of HALO and the other methods as measured by AUROC. We can see that performing vanilla outlier exposure (OE), while excelling in clean OOD detection, does not produce meaningful robustness to detection attacks. ATOM also performs strongly in the clean case and shows some robustness to OOD$\rightarrow$ID attacks, but fails in settings where the ID data is attacked. TRADES is able to achieve some robustness to ID and OOD attacks, especially in the near-OOD datasets (CIFAR-100 and TinyImageNet), but its clean AUROC is weaker than other methods, reducing its practical usefulness. ALOE, OSAD and ATD are generally able to improve clean AUROC (especially for far-OOD) relative to TRADES while retaining reasonable robustness to detection attacks. Comparing with our method, we can see that HALO surpasses existing methods in robust detection performance across all datasets apart from MNIST, where ATD exhibits very strong performance. We hypothesise that the unstable GAN-style training used by ATD produces a detector that is over-fit to this relatively simple dataset, accounting for the extremely high AUROC values. Despite this, we include MNIST results for completeness and note that HALO \textit{significantly outperforms all other robust methods} on this dataset.

\subsection{Sensitivity Analysis} \label{sec:sens}

As HALO has a number of hyperparameters that must be set, in this section we aim to quantitatively understand the effect of various possible choices. To this end, we conduct a sensitivity analysis on the various hyperparameters of our model: $\eta$, $\gamma$, $\beta_1$ and $\beta_2$. In our primary analysis we always set $\beta_1 = \beta_2 = \beta$, with default values of $\eta=2.0$, $\gamma=0.5$ and $\beta = 3.0$, varying only the hyperparameter under examination. To reduce the computation requirements, robust accuracies are evaluated on a subset of the full test set, using AutoAttack's APGD-DLR adversary.\footnote{Empirically this results in a roughly 1\% performance improvement relative to the full AutoAttack benchmark.} Figure \ref{fig:sens} shows average AUROC across the six OOD datasets as well as clean/robust accuracy for various hyperparameter settings. 

Our analysis shows that higher $\eta$ values correspond to improved AUROC across all settings, though this improvement comes at the cost of slightly reduced clean and robust classification accuracy. This follows as larger $\eta$ values more strongly incentivise entropic distributions on OE samples. 

Performance differences over $\gamma$ follow less of a pattern, though we observe that setting $\gamma=0.5$ gives the best robust AUROC and classification accuracy, echoing the results of \citet{rade2022reducing} who introduced the helper term. 

As reported by \citet{zhang2019theoretically}, $\beta$ has a large influence on the clean/robust accuracy trade off, with larger values placing greater emphasis on robustness. The high clean accuracy resultant from lower $\beta$ is accompanied by a high clean AUROC, albeit at the cost of lower AUROC in the `Both' attack setting. Interestingly, we observe that although setting $\beta=6.0$ achieves greater robustness than $\beta=3.0$ under combined attacks, it performs comparatively worse under either single attack, likely due to its lower clean AUROC starting point. 

\begin{figure*}[h!]
\centering
\includegraphics[width=0.97 \textwidth]{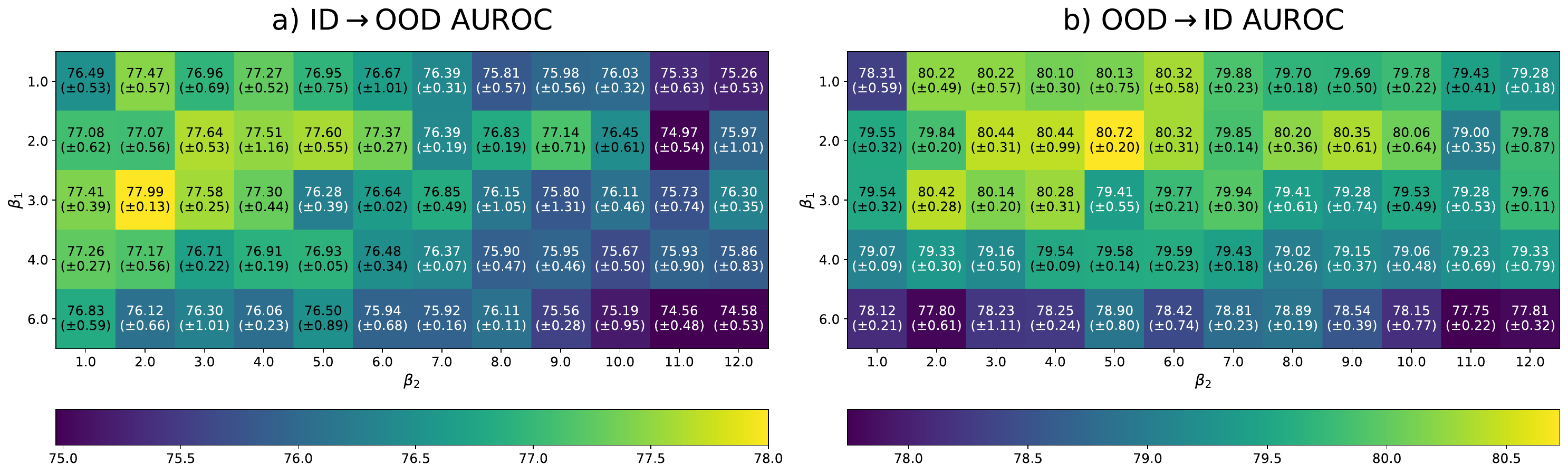}
\caption{AUROC scores over various values of $\beta_1$ and $\beta_2$, under both ID$\rightarrow$OOD and OOD$\rightarrow$ID attacks. We report the average over 3 runs, with the standard deviation in brackets. Different relative values of each hyperparameter lead to different robustness profiles. \vspace{-0.5em}}
\label{fig:doublebeta}
\end{figure*}

Given the significant variation observed with different $\beta$ values, we conducted an additional experiment to better understand the effects of independently varying $\beta_1$ and $\beta_2$. In this analysis, we held fixed values of $\eta = 2.0$ and $\gamma = 0.5$ while performing a parameter sweep over a range of $\beta$ values. We report averages over 3 independent runs, and follow the training methodology described in Section \ref{sec:eval}. Figure \ref{fig:doublebeta} shows the average AUROC of each parameter set under both ID$\rightarrow$OOD and OOD$\rightarrow$ID attacks. 

We observe that for both attack types there are distinct optimal hyperparameter regions, outside which performance degrades notably. Interestingly, while ID$\rightarrow$OOD robustness generally benefits from a higher $\beta_1$:$\beta_2$ ratio and OOD$\rightarrow$ID from a higher $\beta_2$:$\beta_1$ ratio, simply increasing the dominant parameter indefinitely does not yield continued improvements. Instead, performance peaks within a specific range and then deteriorates at more extreme values. This suggests that while the $\beta$ parameters can be tuned to favour particular types of robustness, maintaining balance is important for overall performance. We hypothesise that extreme hyperparameter settings disrupt the joint learning of both clean ID accuracy and robustness, which are both necessary for robust out-of-distribution detection (as discussed in Section \ref{sec:helper_explanation}). On this view, when one parameter becomes too large, the optimisation process disproportionately favours one objective, which degrades overall performance.

For completeness, we present in Table \ref{tab:doublebeta} a performance comparison of three model configurations: the best parameters for ID robustness ($\beta_1 = 3.0$, $\beta_2 = 2.0$), OOD robustness ($\beta_1 = 2.0$, $\beta_2 = 5.0$), and a baseline configuration ($\beta_1 = \beta_2 = 3.0$). We again report averages across 3 independent runs, denoting the models A, B and Ref, respectively. We observe that for almost all datasets, model A outperforms model B in the ID$\rightarrow$OOD direction, and the opposite is true for the OOD$\rightarrow$ID direction. Additionally, for most datasets and on average, the tuned models outperform the reference model in their respective robustness direction. 

Our analysis shows that these hyperparameters can be used to effectively tailor HALO's performance to meet specific application requirements, though care must be taken to ensure a balanced optimisation problem.

\begin{table}[htbp]
\centering
\small
\caption{Comparative analysis of AUROC under different attacks for models A ($\beta_1 = 3.0$, $\beta_2 = 2.0$), B ($\beta_1 = 2.0$, $\beta_2 = 5.0$) and Ref ($\beta_1 = \beta_2 = 3.0$). All results are averaged over 3 independent runs, and best results are \textbf{bolded}.}
\begin{tabular}{lcccc}
\hline
\textbf{Dataset} & \textbf{Direction} & \textbf{A} & \textbf{Ref} & \textbf{B} \\
\hline
\multirow{2}{*}{CIFAR100} & ID$\rightarrow$OOD & 67.52 & \textbf{67.61} & 67.60 \\
& OOD$\rightarrow$ID & 68.24 & 68.51 & \textbf{68.96} \\
\hline
\multirow{2}{*}{TIN} & ID$\rightarrow$OOD & \textbf{77.06} & 76.93 & 76.91 \\
& OOD$\rightarrow$ID & 77.31 & 77.41 & \textbf{77.90} \\
\hline
\multirow{2}{*}{SVHN} & ID$\rightarrow$OOD & \textbf{81.98} & 81.12 & 80.52 \\
& OOD$\rightarrow$ID & \textbf{84.01} & 83.00 & 83.52 \\
\hline
\multirow{2}{*}{Texture} & ID$\rightarrow$OOD & 85.98 & \textbf{86.22} & 85.58 \\
& OOD$\rightarrow$ID & 88.17 & 88.23 & \textbf{88.24} \\
\hline
\multirow{2}{*}{Places365} & ID$\rightarrow$OOD & \textbf{75.88} & 75.61 & 75.57 \\
& OOD$\rightarrow$ID & 75.41 & 75.48 & \textbf{76.05} \\
\hline
\multirow{2}{*}{MNIST} & ID$\rightarrow$OOD & \textbf{79.54} & 78.00 & 79.43 \\
& OOD$\rightarrow$ID & 89.38 & 88.23 & \textbf{89.68} \\
\hline
\multirow{2}{*}{Average} & ID$\rightarrow$OOD & \textbf{77.99} & 77.58 & 77.60 \\
& OOD$\rightarrow$ID & 80.42 & 80.14 & \textbf{80.73} \\
\hline
\end{tabular}
\label{tab:doublebeta}
\end{table}

\subsection{Different Attack Parameters} \label{sec:attack_params}

\begin{figure*}[h!]
\centering
\includegraphics[width=0.92\textwidth]{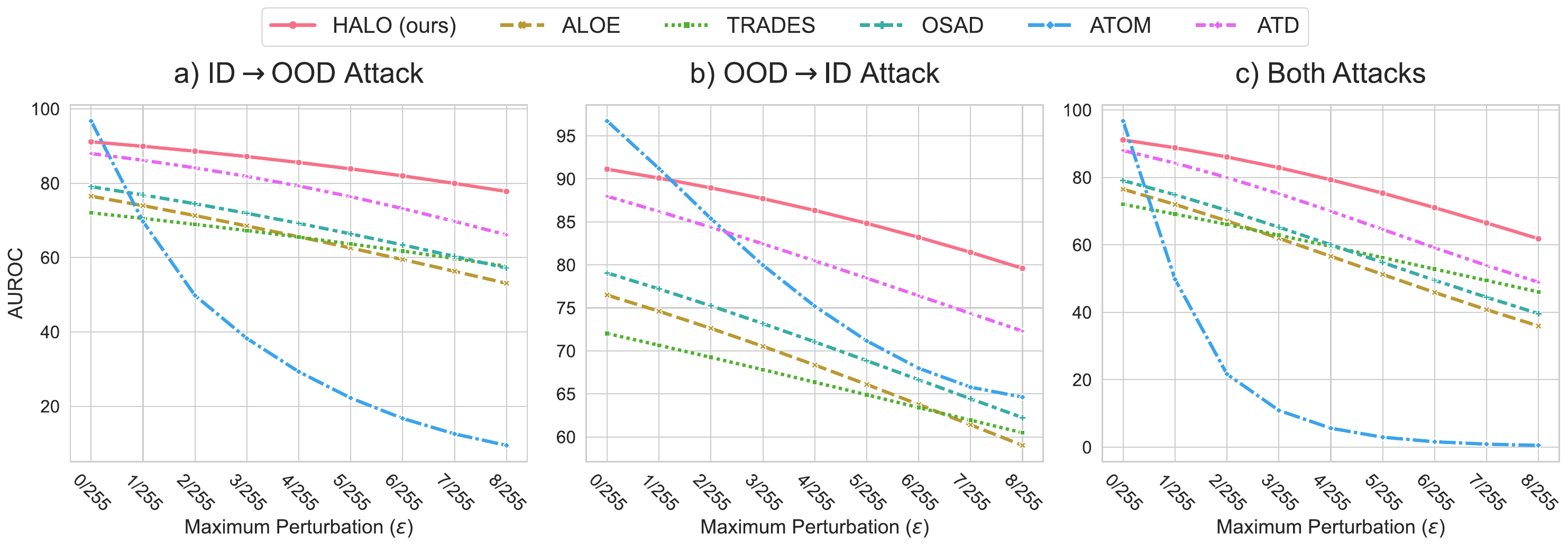}
\caption{Comparison of robust OOD detection methods across multiple attack strengths. Subplots left to right: a) Only ID data is attacked, b) Only OOD data is attacked, c) Both ID and OOD data attacked. Our method, HALO, outperforms existing approaches for all reasonable attack strengths, across all threat models.}
\label{fig:eps_plot}
\end{figure*}

To better understand HALO and other methods' robustness to detection attacks, we perform tests with a varied maximum perturbation budget or number of PGD steps. For these experiments, we use $\epsilon=8/255$ and 40 PGD steps by default, varying only the parameter in question. Figure \ref{fig:eps_plot} shows the robust OOD detection performance of each method over a range of maximum perturbations. We observe that HALO has the strongest performance across all attack settings and all budgets, except for an OOD$\rightarrow$ID attack with $\epsilon=1/255$, where ATOM is slightly better. For all realistic budgets, HALO is significantly better. 
In Table \ref{tab:auroc_attack_steps_comparison} we show the average AUROC in different attack settings for PGD attacks with a varied number of steps. Performance in all settings appears to plateau at around 40 steps, justifying this choice for our evaluations. 
\begin{table}[t!]
\centering
\caption{Comparison of average AUROC for different number of attack steps. For all attacks, $\epsilon=8/255$ and $\alpha=2.5\times\frac{\epsilon}{\text{n\_steps}}$. Attack strength appears to plateau at around 40 steps.}
\begin{tabular}{cccc}
\toprule
\textbf{\# Steps} & \textbf{OOD→ID} & \textbf{ID→OOD} & \textbf{Both} \\
\midrule
5 & 80.15 & 78.35 & 63.06 \\
10 & 79.78 & 77.96 & 62.23 \\
20 & 79.65 & 77.83 & 61.94 \\
40 & 79.59 & 77.78 & 61.81 \\
80 & 79.57 & 77.76 & 61.76 \\
120 & 79.56 & 77.75 & 61.75 \\
\bottomrule
\end{tabular}
\label{tab:auroc_attack_steps_comparison}
\end{table}

\subsection{Black Box Attacks} \label{sec:tfr}
To validate that HALO's robustness is not merely an artifact of gradient masking \cite{athalye2018obfuscated}, we evaluate its performance against two types of black-box attacks: transfer attacks and gradient-free attacks. Transfer attacks, where adversarial examples crafted for one model are used to attack another \cite{papernot2016transferability}, serve as an important diagnostic tool for detecting gradient obfuscation. We evaluate HALO's robustness against transfer attacks by exchanging adversarial examples with ATD, another leading approach in robust OOD detection. To complement this experiment, we also evaluate HALO against an adaptation of Square Attack \cite{andriushchenko2020square}, a state-of-the-art gradient-free method. Our results show that HALO maintains strong robust performance under these attacks and that they are both somewhat less effective than white-box PGD approaches. We direct the reader to Appendix \ref{apdx:blackbox_details} for complete experimental details, results, and analysis of both attack types. 

\subsection{Performance with Extra Data} \label{sec:extra_data}
\begin{table*}[t!]
\centering
\caption{Comparison of ATD and HALO models trained with extra pseudo-labelled ID data, best values in \textbf{bold}. Both models are trained with CIFAR-10 as the ID dataset with extra data sourced in the same manner as \citet{carmon2019unlabeled}. A PreAct ResNet-18 \cite{he2016identity} is trained for 400 epochs with 70\% of the batch comprising pseudo-labelled data. \vspace{0.75em}}

\begin{tabular}{lcccccccc}
\toprule
\multirow{2}{*}{Dataset} & \multicolumn{4}{c}{\textbf{ATD}} & \multicolumn{4}{c}{\textbf{HALO}} \\
\cmidrule(lr){2-5} \cmidrule(lr){6-9}
 & Clean & ID→OOD & OOD→ID & Both & Clean & ID→OOD & OOD→ID & Both \\
\midrule
CIFAR100 & 84.21 & 63.56 & 68.11 & 44.66 & \textbf{86.88} & \textbf{69.31} & \textbf{69.06} & \textbf{49.39} \\
TIN      & 90.00 & 71.21 & \textbf{76.39} & 52.30 & \textbf{93.14} & \textbf{77.93} & 76.36 & \textbf{55.58} \\
SVHN     & 92.08 & 73.82 & 79.84 & 54.21 & \textbf{97.51} & \textbf{87.72} & \textbf{85.50} & \textbf{64.60} \\
Texture  & 91.03 & 73.67 & 79.03 & 55.89 & \textbf{98.13} & \textbf{90.00} & \textbf{89.85} & \textbf{72.67} \\
Places365& 90.22 & 71.25 & \textbf{76.06} & 51.69 & \textbf{92.39} & \textbf{76.92} & 74.14 & \textbf{53.23} \\
MNIST    & \textbf{99.43} & \textbf{97.24} & \textbf{99.09} & \textbf{95.92} & 91.48 & 77.92 & 85.55 & 69.40 \\
\bottomrule
\end{tabular}
\label{tab:extra-data}
\end{table*}
Unlike other methods, ATD uses a pre-trained robust model to produce a feature vector for each input that is fed into a discriminator, producing an OOD score for the detection task. Importantly, the authors evaluate their method using a robust extractor model that was trained using extra, pseudo-labelled data as in \citet{carmon2019unlabeled}. This method involves taking a large dataset (80M Tiny Images \cite{torralba200880} in this case) finding a subset of that dataset that a standard classifier has high confidence on and assigning it pseudo-labels based on the predicted class. \citet{carmon2019unlabeled} show that this method considerably increases both clean and robust accuracy. 

As the training process of ATD implicitly relies on the availability of extra data that other methods do not have access to, comparison between it and other methods is potentially biased. For this reason, in our testing of ATD we use a feature extractor model that is trained without extra data.

However, the question of how these methods perform when given extra data remains. To compare ATD and HALO in the extra data setting, we train a PreAct ResNet-18 \cite{he2016identity} for 400 epochs using a similar training and evaluation method outlined in Section \ref{sec:eval}. The full training method is detailed in Appendix \ref{apdx:extra_data_details}. We show in Table \ref{tab:extra-data} that when both methods have access to extra data, HALO retains its advantage in almost all settings, though performance is closer than in the original setting. We hypothesise that this is in part due to HALO having stronger performance to begin with, and marginal improvements becoming more difficult as performance improves.

\subsection{Post-processor Study} \label{sec:ppstudy}

An advantage of HALO and other training-based OOD detection methods is that they are compatible with much of the vast array of post-processing techniques that have been developed for OOD detection. Here, we perform a series of experiments, evaluating the detection performance of HALO when paired with a number of post-processors. Results of these experiments are shown in Table \ref{tab:postprocessor_results}. Looking at the baseline method of maximum softmax probability (MSP) \cite{hendrycks17baseline}, it is clear that there is room to improve. Methods such as ODIN \cite{liang2018enhancing}, OpenMax \cite{bendale2016openmax}, Energy \cite{liu2020energy} and Generalised Entopy (GEN) \cite{liu2023gen} all improve upon this baseline, showing a similar performance profile across the different attack settings. We also observe that some methods that report stronger performance than the baseline on standard (non-robust) models (Simplified Hopfield Energy \cite{zhang2022out} and RMD \cite{ren2021simple}) see reduced effectiveness when paired with our adversarially trained model. This is perhaps indicative of some internal structural differences between clean and robust models that are important to the performance of these methods. These results suggest that unlike other methods that prescribe a specific detection mechanism, HALO can be paired with new post-processors as they are developed, potentially increasing performance down the line.

\begin{table}[t]
\centering
\small
\caption{Clean and attacked AUROC of HALO when paired with different choices of post-processor. CIFAR-10 is used as the ID dataset and average AUROCs are reported over OpenOOD datasets. The baseline method, MSP \cite{hendrycks17baseline}, is in \textit{italics} and highest results are in \textbf{bold}. More sophisticated methods such as Energy-based OODD \cite{liu2020energy} and Generalised ENtropy \cite{liu2023gen} perform noticeably better than the baseline method.}
\begin{tabular}{lcccc}
\toprule
\multirow{2}{*}{\textbf{Model}} & \multicolumn{4}{c}{\textbf{Avg. AUROC}} \\
\cmidrule(lr){2-5}
 & \textbf{Clean} & \textbf{ID→OOD} & \textbf{OOD→ID} & \textbf{Both} \\
\midrule
HALO + RMDS & 83.23 & 69.65 & 74.39 & 59.61 \\
HALO + SHE & 85.79 & 74.49 & 74.98 & 60.80 \\
\textit{HALO + MSP} & \textit{89.87} & \textit{76.80} & \textit{78.52} & \textit{61.82} \\
HALO + ODIN & 90.54 & 77.72 & 79.17 & 62.42 \\
HALO + OpenMax & 90.64 & 77.97 & 79.48 & 62.49 \\
HALO + Energy & 91.12 & \textbf{78.17} & \textbf{79.96} & \textbf{62.66} \\
\textbf{HALO + GEN} & \textbf{91.13} & \textbf{78.17} & \textbf{79.96} & \textbf{62.66} \\
\bottomrule
\end{tabular}
\label{tab:postprocessor_results}
\end{table}

\section{Conclusion} \label{sec:conc}

This work addresses the challenge of robust out-of-distribution detection in machine learning models. We introduce HALO, a novel approach that jointly optimises for classification performance, OOD detection, and adversarial robustness. Through comprehensive experiments, we demonstrate that HALO achieves state-of-the-art performance in both clean and adversarial settings across various datasets, outperforming existing methods, particularly when both ID and OOD samples are subject to attack. Our sensitivity analyses show the effect of each hyperparameter, and our black box attack experiments suggest HALO's robustness is strong against varied attacks. HALO's theoretical framework and empirical advances contribute to our understanding of how to build more trustworthy AI systems that can reliably identify out-of-distribution inputs, even under adversarial attack. \\
\indent\textbf{Limitations:} An important finding from our study is that current robust OOD detection methods, including HALO, face significant challenges when scaling to larger, more complex datasets, limiting their usefulness. We show in Appendix \ref{apdx:extra_results} the performance of various methods when CIFAR-100 and TinyImageNet are used as the ID dataset. The results show that although HALO maintains its advantage over other methods in these settings, performance across the board is significantly degraded compared to simpler datasets. For example, when CIFAR-100 is used as the ID dataset, no method achieves an AUROC above 50 in the setting where both attacks are performed, and most methods struggle to maintain reasonable performance even under single-sided attacks. This performance drop on larger datasets appears to be influenced by factors beyond simply the number of classes (see Appendix \ref{apdx:class_cardinality}), suggesting deeper underlying challenges in robust OOD detection at scale. \\
\indent\textbf{Future Work:} While our work represents a significant step towards robust OOD detection, several important research directions remain. Future work should focus primarily on understanding and addressing the fundamental challenges of scaling robust OOD detection to larger, more complex datasets. This includes investigating the root causes of performance degradation, such as the roles of feature space complexity and inter-class similarity. Additionally, future work could expose OOD detection methods to new attack types and threat models, as well as explore robust OOD detection in domains beyond image classification.

\section*{Acknowledgments}
This research was undertaken using the LIEF HPC-GPGPU Facility hosted at the University of Melbourne. 
This Facility was established with the assistance of LIEF Grant LE170100200. 
Sarah Erfani is in part supported by Australian Research Council~(ARC) Discovery Early Career Researcher Award~(DECRA) DE220100680. Christopher Leckie is in part supported by the ARC Centre of Excellence on Automated Decision Making and Society CE200100005.

\bibliographystyle{IEEEtranN}
\bibliography{references}

\clearpage
\newpage


\appendices
\begin{center}
\Large{HALO: Robust Out-of-Distribution Detection via Joint Optimisation}
\end{center}

\section{Real-World Attack Scenarios} \label{apdx:real_world}

To better understand the practical implications of different OOD detection attacks, we consider their analogues in the autonomous driving scenario discussed in Section \ref{sec:intro}. This domain is particularly instructive as it involves safety-critical decisions where OOD detection performance can have significant consequences.

\textbf{ID$\rightarrow$OOD}: An ID$\rightarrow$OOD attack in this context might involve an adversary subtly modifying the appearance of a normal road scene to make it appear anomalous to the system. For instance, perturbations could cause the system to interpret a standard driving environment with good visibility as potentially hazardous, triggering unnecessary defensive actions such as speed reduction or route modification. Such attacks could significantly impair system efficiency and reliability.

\textbf{OOD$\rightarrow$ID}: Conversely, an OOD$\rightarrow$ID attack could potentially cause the system to misinterpret genuinely anomalous conditions as normal operating scenarios. An example would be the modification of visual inputs during adverse weather conditions, such as dense fog or heavy snow, to appear as standard driving conditions to the detection system. This type of attack could lead the system to maintain normal operation when increased caution is warranted, exposing the passengers to risk.

\textbf{Both Attacks}: The scenario where both attack types are perpetrated is more speculative but warrants consideration nonetheless. An adversary capable of both making safe conditions appear dangerous and dangerous conditions appear safe would have significantly greater capacity to compromise the utility of an OOD detection system, making it's outputs difficult to trust. 

Beyond autonomous driving, the implications of robust OOD detection extend to numerous safety-critical domains. Consider, for instance, medical diagnostic systems where reliable identification of unusual cases could be crucial for patient safety, or content moderation systems where accurate detection of novel forms of harmful content is essential for platform integrity. These scenarios further underscore the fundamental importance of developing detection methods that maintain their reliability even in the presence of adversarial attacks.

\section{Methods of Detection Attack} \label{apdx:detection_attacks}

Here we give a detailed description of the method of attack for different OOD detection methods.

\subsection{Output Distribution Based Methods}

For methods that use the softmax distribution to calculate an OOD score e.g. MSP, we aim to move the distribution closer to or further away from the uniform distribution, depending on the type of attack. In Algorithm \ref{alg:logits} we provide pseudo-code to produce an adversarial example with this method.

\begin{algorithm}[!h]
\caption{Generate Adversarial Examples (Logits)}
\begin{algorithmic}[1]
\Require Model $f$, input $X$, perturbation size $\epsilon$, number of steps $N$, step size $\alpha$, attack type $T$
\State $\mathbf{X}_{adv} \gets \mathbf{X} + \mathcal{U}(-\epsilon, \epsilon)$
\For{$t = 1$ to $N$}
\State $\mathbf{z} \gets f(\mathbf{X}_{adv})$
\If{$T$ is ``OOD$\rightarrow$ID''}
\State $\mathcal{L} \gets -\bigl(\frac{1}{K}\sum_{k=1}^K z_k - \log\sum_{k=1}^K e^{z_k}\bigr)$ \Comment{Minimise entropy}
\ElsIf{$T$ is ``ID$\rightarrow$OOD''}
\State $\mathcal{L} \gets \frac{1}{K}\sum_{k=1}^K z_k - \log\sum_{k=1}^K e^{z_k}$ \Comment{Maximise entropy}
\EndIf
\State $\nabla \gets \nabla_{\mathbf{X}_{adv}} \mathcal{L}$
\State $\mathbf{X}_{adv} \gets \mathbf{X}_{adv} + \alpha \cdot \text{sign}(\nabla)$
\State $\mathbf{X}_{adv} \gets \text{clip}(\mathbf{X}_{adv}, \mathbf{X} - \epsilon, \mathbf{X} + \epsilon)$
\State $\mathbf{X}_{adv} \gets \text{clip}(\mathbf{X}_{adv}, 0, 1)$
\EndFor
\State \Return $\mathbf{X}_{adv}$
\end{algorithmic}
\label{alg:logits}
\end{algorithm}

We note that this method is effective when the detection method relies on the relative uniformity or non-uniformity of the softmax distribution (MSP, GEN, ODIN) but needs to be modified for other methods such as Relative Mahalanobis Distance (RMD) which do not derive scores from uniformity. For this reason, attacks targeting such methods are potentially unreliable and may over-estimate robustness. In our case, as this non-optimal attack method is able to significantly reduce AUROC for RMD, we do not test with the modified, optimal attack, having already demonstrated the vulnerability of this OOD detection method. 


\subsection{Adversarially Trained Discriminator (ATD)}

As ATD uses a discriminator to produce an OOD score between 0 and 1 (with 1 indicating ID and 0 indicating OOD), we can simply use the binary cross entropy loss between the discriminator's output and the true distribution of the data point to produce attacks \cite{azizmalayeri2022your}. Algorithm \ref{alg:atd} shows pseudo-code that performs this attack:

\begin{algorithm}[!h]
\caption{Generate Adversarial Examples (ATD)}
\begin{algorithmic}[1]
\Require Feature extractor model $f$, discriminator $D$, input $X$, perturbation size $\epsilon$, number of steps $N$, step size $\alpha$, attack type $T$
\State $\mathbf{X}_{adv} \gets \mathbf{X} + \mathcal{U}(-\epsilon, \epsilon)$
\For{$t = 1$ to $N$}
\State $\mathbf{z} \gets D(f(\mathbf{X}_{adv}))$
\If{$T$ is ``OOD$\rightarrow$ID''}
\State $\mathbf{y} \gets \mathbf{0}$ 
\ElsIf{$T$ is ``ID$\rightarrow$OOD''}
\State $\mathbf{y} \gets \mathbf{1}$
\EndIf
\State $\mathcal{L} \gets \text{BCE}(\mathbf{z},\mathbf{y}) $ 
\State $\nabla \gets \nabla_{\mathbf{X}_{adv}} \mathcal{L}$
\State $\mathbf{X}_{adv} \gets \mathbf{X}_{adv} + \alpha \cdot \text{sign}(\nabla)$
\State $\mathbf{X}_{adv} \gets \text{clip}(\mathbf{X}_{adv}, \mathbf{X} - \epsilon, \mathbf{X} + \epsilon)$
\State $\mathbf{X}_{adv} \gets \text{clip}(\mathbf{X}_{adv}, 0, 1)$
\EndFor
\State \Return $\mathbf{X}_{adv}$
\end{algorithmic}
\label{alg:atd}
\end{algorithm}


\subsection{Adversarial Training with Informative Outlier Mining (ATOM)}

The core component of ATOM is a $K+1$ way classifier, where the first $K$ classes are ID classes and the $K+1$'th class represents OOD input. As such, we must adapt our detection attack to target increased/decreased probability mass on the $K+1$'th class. Algorithm \ref{alg:atom} shows pseudo-code to perform this attack:

\begin{algorithm}[!h]
\caption{Generate Adversarial Examples (ATOM)}
\begin{algorithmic}[1]
\Require $K+1$ way classification model $f$, input $X$, perturbation size $\epsilon$, number of steps $N$, step size $\alpha$, attack type $T$
\State $\mathbf{X}_{adv} \gets \mathbf{X} + \mathcal{U}(-\epsilon, \epsilon)$
\State $\mathbf{y} \gets \mathbf{e}_{K+1}$ \Comment{Standard basis vector with 1 in K+1th position}
\For{$t = 1$ to $N$}
\State $\mathbf{z} \gets D(f(\mathbf{X}_{adv}))$
\If{$T$ is ``OOD$\rightarrow$ID''}
\State $\mathcal{L} \gets \text{CE}(\mathbf{z}, \mathbf{y})$
\ElsIf{$T$ is ``ID$\rightarrow$OOD''}
\State $\mathcal{L} \gets -\text{CE}(\mathbf{z}, \mathbf{y})$
\EndIf
\State $\nabla \gets \nabla_{\mathbf{X}_{adv}} \mathcal{L}$
\State $\mathbf{X}_{adv} \gets \mathbf{X}_{adv} + \alpha \cdot \text{sign}(\nabla)$
\State $\mathbf{X}_{adv} \gets \text{clip}(\mathbf{X}_{adv}, \mathbf{X} - \epsilon, \mathbf{X} + \epsilon)$
\State $\mathbf{X}_{adv} \gets \text{clip}(\mathbf{X}_{adv}, 0, 1)$
\EndFor
\State \Return $\mathbf{X}_{adv}$
\end{algorithmic}
\label{alg:atom}
\end{algorithm}

\section{Proofs} \label{apdx:proofs}

Here we give a proof that minimising $\phi(g'(\mathbf{X})Y_{det})$ is equivalent to minimising KL($f_\theta(\mathbf{X}), \mathcal{U}$) for OE samples ($Y_{det} = 1$). 

We start with the definitions and assumptions:

Let $\phi(\alpha) = \exp(-\alpha)$ be our loss function.
Let $\bm{x}$ be an OE sample and $y_{det} = 1$ be the detection label, indicating that it is from the OE distribution.
Assume $g(\bm{x}) = H(f(\bm{x}))$, where $H$ is the Shannon entropy and $f(\bm{x})$ is our classifier output. We further define $g'(\bm{x}) = g(\bm{x}) - \tau$, where $g(\bm{x})$ is our OOD detector and $\tau$ is the detection threshold.

Our goal is to show that minimising $\phi(g'(\bm{x})y_{det})$ for OE data is equivalent to minimising $\text{KL}(f(\bm{x}), \mathcal{U})$, where $\mathcal{U}$ is the uniform distribution.

For OE data, $y_{\text{det}} = 1$. Thus, we want to minimise:
   $$\phi(g'(\bm{x})y_{\text{det}}) = \phi(g'(\bm{x})) = \exp(-g'(\bm{x}))$$
Expanding $g'(\bm{x})$:
   $$\exp(-g'(\bm{x})) = \exp(-(g(\bm{x}) - \tau)) = \exp(-g(\bm{x})) \cdot \exp(\tau)$$

Since $\exp(\tau)$ is a constant, minimising $\exp(-g'(\bm{x}))$ is equivalent to minimising $\exp(-g(\bm{x}))$.

Recall that $g(\bm{x}) = H(f(\bm{x}))$. Thus, we are minimising:
   $$\exp(-H(f(\bm{x})))$$

Since the exponential function is monotonically increasing, minimising $\exp(-H(f(\bm{x})))$ is equivalent to maximising $H(f(\bm{x}))$.

Now, consider the KL divergence between $f(\bm{x})$ and the uniform distribution $\mathcal{U}$:
   \begin{align*}
   \text{KL}(f(\bm{x}), \mathcal{U}) &= \sum_{i=1}^K f_i(\bm{x}) \log \frac{f_i(\bm{x})}{1/K} \\
   &= \log K + \sum_{i=1}^K f_i(\bm{x}) \log f_i(\bm{x})
   \end{align*}
   where $K$ is the number of classes.

Recall the definition of entropy:
   $$H(f(\bm{x})) = -\sum_{i=1}^K f_i(\bm{x}) \log f_i(\bm{x})$$
Substituting this into the KL divergence expression:
   $$\text{KL}(f(\bm{x}), \mathcal{U}) = \log K - H(f(\bm{x}))$$

Since $\log K$ is constant for a given problem, minimising $\text{KL}(f(\bm{x}), \mathcal{U})$ is equivalent to maximising $H(f(\bm{x}))$.

\section{Toy Model Details} \label{apdx:toy_model} 

Here we explain in detail the toy model that we study in Section \ref{sec:toy}. 

\begin{figure}[h!]
\centering
\includegraphics[width=0.3\textwidth]{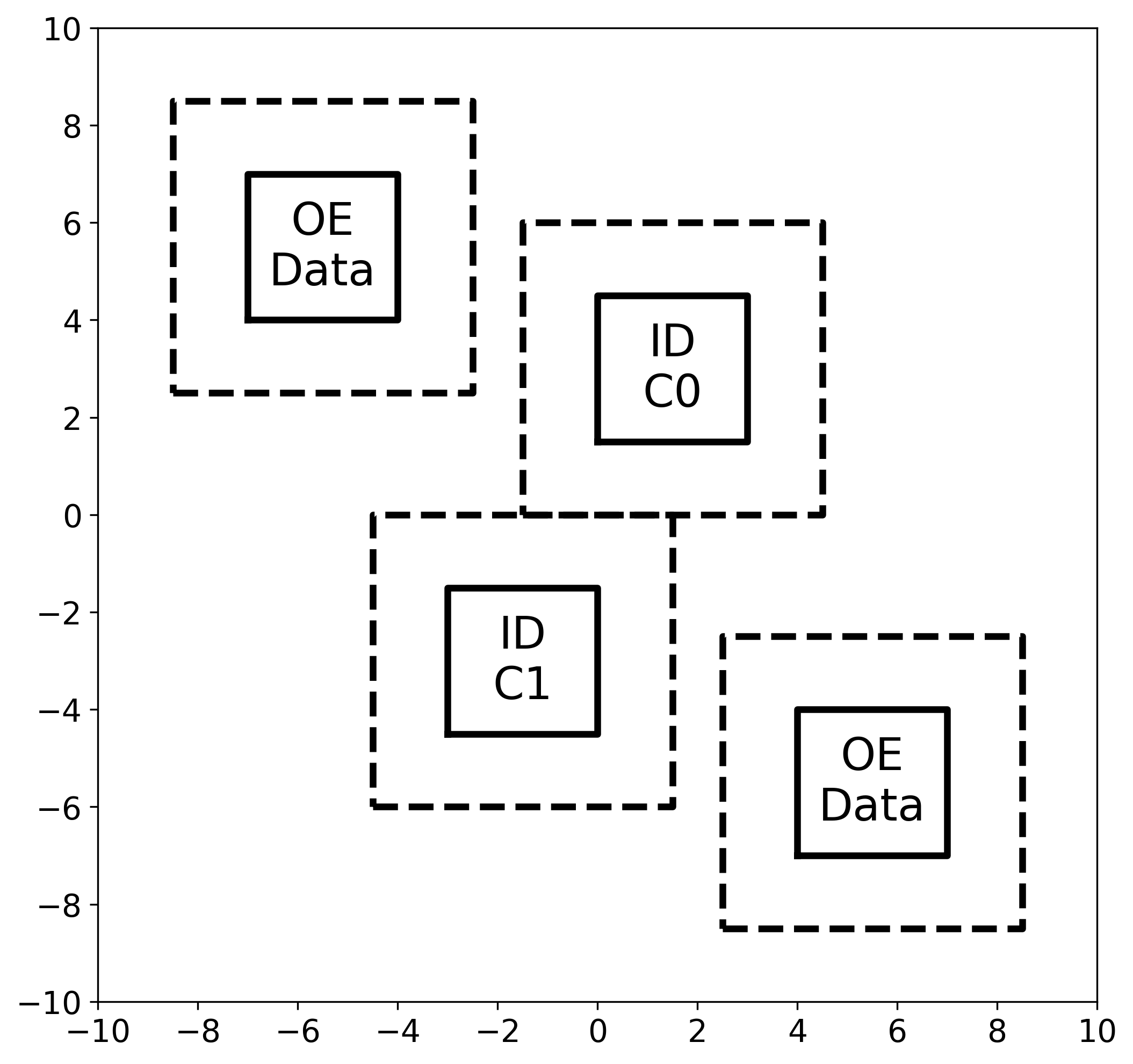}
\caption{Toy model data distributions. Solid lines represent data distributions and dashed lines represent allowable perturbations.}
\label{fig:toy_classes}
\end{figure}

We aim to demonstrate the effects of different adversarial training approaches in terms of what kinds of robustness they create in models. We consider a simple binary classification task in a 2D space spanning -10 to 10 units in both x and y dimensions. There are 3 data distributions for this task which are visualised in Figure \ref{fig:toy_classes}:
\begin{itemize}
    \item ID class 0: $\{(x, y) \in \mathbb{R}^2 \mid x \in [0, 3], y \in [1.5, 4.5]\}$
    \item ID class 1: $\{(x, y) \in \mathbb{R}^2 \mid x \in [-3, 0], y \in [-4.5, -1.5]\}$
    \item OE Data: $\{(x, y) \in \mathbb{R}^2 \mid (x \in [-7, -4] \wedge y \in [4, 7]) \vee (x \in [4, 7] \wedge y \in [-7, -4])\}$
\end{itemize}

When training, we randomly sample 1000 points from each of these distributions to serve as the ID and OE sets.

We use the $\ell_\infty$ threat model and set the maximum allowable perturbation magnitude to be $\epsilon=1.5$. We note that based on these data distributions and perturbation strength, it is possible to construct a model with perfect clean \textbf{and} robust performance in both the ID classification and OOD detection tasks.

For our model, we choose a simple MLP with two hidden layers, both of size 64. We train for 400 epochs using SGD with a learning rate of 0.001. To highlight the inadequacies of certain training methods we train four different models, each with a different objective function. All methods are variations on the following objective function:
\begin{equation*}
    \begin{aligned}
        \mathbb{E}_{(\bm{x},y)\sim\mathcal{D}_{\text{in}}} \bigl[
        \underbrace{\text{CE}(f_\theta(\bm{x}), y)}_{\text{a) Clean accuracy term}} + \max_{\bm{\delta} \in \mathcal{S}} \underbrace{\text{KL} (f_\theta(\bm{x}) , f_\theta(\bm{x} +
        \bm{\delta}))}_{\text{b) ID robustness term}}\bigr] + \\ \mathbb{E}_{\bm{x}' \sim \mathcal{D}_{\text{oe}}} \bigl[\underbrace{\text{KL}(f_\theta(\bm{x}'), \mathcal{U})}_{\text{c) OE term}} + \max_{\bm{\delta} \in \mathcal{S}} \underbrace{\text{KL} (f_\theta(\bm{x}') , f_\theta(\bm{x}' + \bm{\delta}))}_{\text{d) OOD robustness term}}
        \bigr]
    \end{aligned}
\end{equation*}

The standard model is trained using only terms (a) and (c), whereas the ID attacked and OOD attacked models also use terms (b) and (d), respectively. The model labelled `both' uses all four terms such that both ID and OE datasets are attacked.

\section{Training Setup and Evaluation Methodology}

\subsection{Training Setup}\label{apdx:trainingsetup}

In Section \ref{sec:eval} we test HALO on real-world datasets, using ResNet models as our base classifier. Here we review the training setup we use in detail, which mostly identical to that used in \citet{rade2022reducing}.

\textbf{Standard Setting:} We train a ResNet-18 \cite{he2016deep} for 50 epochs, using adversarial validation to select the best model. We use SGD with Nesterov momentum \cite{nesterov1983method}, momentum factor 0.9 and weight decay 0.0005. Learning rates follow a single cycle schedule with cosine annealing \cite{smith2019super} and a maximum learning rate of 0.21. During training we use a PGD adversary with a maximum $\ell_\infty$ perturbation of $\epsilon= 8/255$ and 10 steps of size $\alpha = 2/255$. The attack aims to maximise the TRADES KL term (2nd term of Equation \ref{eq:trades}.) We use the same attack for adversarial validation. Additionally, we use a batch size of 128 for both ID and OE datasets and apply commonly used data augmentation techniques: random crops with a padding size of 4 and random horizontal flips. 

\textbf{Extra Data Setting:} When using extra data as in Section \ref{sec:extra_data}, we use a PreAct ResNet-18 \cite{he2016identity} and train for 400 epochs. We mostly use the same setup as in the standard setting with the following exceptions: the maximum learning rate is set to 0.4 and we employ model weight averaging \cite{izmailov2018averaging} with decay rate 0.995. We also use a batch size of 1024 with 70\% of the data coming from the pseudo-labelled set. We use extra data provided by \citet{carmon2019unlabeled}.\footnote{Extra data available at: \href{https://github.com/yaircarmon/semisup-adv}{https://github.com/yaircarmon/semisup-adv}}

\textbf{Additional Notes:} Some methods we consider such as HAT and HALO require a helper model. To create this model, we simply use standard (non-robust) training, following the prescribed method for the given setting. For ATD, we use a HAT model trained following the above method as the feature extractor model, and make use of the authors' original code for training the generator and discriminator. Likewise for OSAD and ATOM, we make use of the publicly available training code, matching hyperparameter settings.\footnote{Implementations used for each method:\\ OSAD: \href{https://github.com/rshaojimmy/OSAD}{https://github.com/rshaojimmy/OSAD}\\ ATOM: \href{https://github.com/jfc43/informative-outlier-mining}{https://github.com/jfc43/informative-outlier-mining} \\ ATD: \href{https://github.com/rohban-lab/ATD}{https://github.com/rohban-lab/ATD}}

\subsection{Evaluation Methodology} \label{apdx:eval_setup}

\textbf{Detection Attacks:} For detection attacks, unless otherwise stated we use a PGD adversary with maximum $\ell_\infty$ perturbation $\epsilon=8/255$ and 40 steps of size $\alpha = 0.5/255$. The objective that this adversary is aiming to maximise/minimise depends on the method being attacked. For details of the specific attack implementations, see Appendix \ref{apdx:detection_attacks}. 

By default the OpenOOD datasets are normalised to follow a normal distribution with mean 0 and standard deviation 1. For the purpose of our attacks we standardise these images so that values lie between 0 and 1. We perform our attack in this space, clipping perturbed images to ensure all pixel values stay within this range. After the attack is performed, we reverse the standardisation before performing OOD detection on the attacked samples.

\textbf{Classification Attacks:} We evaluate robust classification accuracy using AutoAttack with the standard combination of four adversaries. 

\section{Black Box Attacks} \label{apdx:blackbox_details}

Here, we detail the black box attacks that we perform on HALO to establish that its robustness is not merely due to gradient masking. 

\subsection{Transfer Attacks}
\begin{table*}[t!]
\small
\centering
\caption{Transferability study comparing HALO and ATD with CIFAR-10 (C-10) and CIFAR-100 (C-100) as ID datasets, and SVHN, CIFAR-100, or CIFAR-10 as OOD datasets. For HALO, the MSP postprocessor \cite{hendrycks17baseline} is used. Differences in effectiveness are reported in brackets. Overall, transferred attacks appear to be moderately effective compared to no attack but are significantly worse than model-specific attacks. There does not appear to be any significant asymmetry in the effectiveness of transferred attacks in the CIFAR-10 setting, however when CIFAR-100 is ID, HALO appears to be more resistant to ATD-crafted perturbations than vice-versa. }
\begin{tabular}{cclccccccc}
\toprule
\multirow{2}{*}{\textbf{ID}} & \multirow{2}{*}{\textbf{OOD}} & \multirow{2}{*}{\textbf{Model}} & \multicolumn{4}{c}{\textbf{Own Attack AUROC}} & \multicolumn{3}{c}{\textbf{Transferred Attack AUROC ($\Delta$)}} \\
\cmidrule(lr){4-7} \cmidrule(lr){8-10}
& & & \textbf{Clean} & \textbf{ID→OOD} & \textbf{OOD→ID} & \textbf{Both} & \textbf{ID→OOD} & \textbf{OOD→ID} & \textbf{Both} \\
\midrule
\multirow{4}{*}{C-10} & \multirow{2}{*}{C-100} & HALO & 85.71 & 65.97 & 67.93 & 45.48 & 78.28 (+12.31) & 78.34 (+10.41) & 69.15 (+23.67) \\
& & ATD & 84.25 & 62.99 & 67.74 & 43.80 & 76.11 (+13.12) & 77.67 (+9.93) & 67.96 (+24.16) \\
\cmidrule{2-10}
& \multirow{2}{*}{SVHN} & HALO & 95.03 & 78.57 & 81.66 & 57.52 & 90.61 (+12.04) & 90.40 (+8.74) & 83.52 (+26.00) \\
& & ATD & 92.19 & 73.37 & 79.66 & 53.32 & 86.42 (+13.05) & 87.27 (+7.61) & 78.95 (+25.63) \\
\midrule
\multirow{4}{*}{C-100} & \multirow{2}{*}{C-10} & HALO & 67.77 & 43.87 & 43.09 & 24.66 & 63.41 (+19.54) & 62.44 (+19.35) & 57.81 (+33.15) \\
& & ATD & 56.97 & 34.58 & 39.70 & 23.42 & 50.35 (+15.77) & 51.74 (+12.04) & 45.11 (+21.69) \\
\cmidrule{2-10}
& \multirow{2}{*}{SVHN} & HALO & 83.59 & 58.69 & 57.91 & 34.66 & 80.59 (+21.90) & 78.36 (+20.45) & 74.89 (+40.23) \\
& & ATD & 78.22 & 56.98 & 56.73 & 34.86 & 73.96 (+16.98) & 68.29 (+11.56) & 63.31 (+28.45) \\
\bottomrule
\end{tabular}
\label{tab:transferability_results}
\end{table*}
In the context of classification robustness, a transfer attack involves crafting perturbed images for a model A and using them to fool a model B \cite{papernot2016transferability}. This approach is often effective in producing misclassification in model B, in some cases even being \textit{more effective} than attacks crafted specifically for model B \cite{demontis2019adversarial}. One hypothesis for this phenomena is that model B may have internalised a shallow robustness that works via \textit{gradient obfuscation} \cite{athalye2018obfuscated}, making gradient-based attacks ineffective. For this reason, transfer attacks are an important way of interrogating the true degree of a model's robustness. 
Therefore, to ensure that our model's robustness is not simply the result of an under-powered attack, we conduct a series of transferred \textit{detection} attacks between HALO and another strong approach: ATD. Table \ref{tab:transferability_results} shows the results of this experiment. We consider two choices for the ID dataset: CIFAR-10 and CIFAR-100 and use SVHN and CIFAR-100/CIFAR-10 as the OOD datasets, respectively. We use the ATD model trained with extra data for CIFAR-10, to more closely match to HALO's performance. For HALO, we use the MSP post-processor. We find that transferred attacks are somewhat effective, but significantly worse than model-specific attacks. When using CIFAR-10 as the ID dataset, there appears to be no asymmetry in the effectiveness of transferred attacks, with both models performing better by an approximately equal margin on transferred attacks. In contrast, for CIFAR-100, we observe that there is a noticeable asymmetry in effectiveness, where HALO appears more resistant to transferred attacks. 

\subsection{Square Attack}
\begin{table}[t]
\small
\centering
\caption{Results of Square Attack targeting detection when CIFAR-10 is the ID dataset and CIFAR-100/SVHN are the OOD datasets. 1000 iterations are used to craft each attack. Square attack tends to perform slightly worse than white-box attacks, and similarly to transferred attacks. }
\begin{tabular}{lcccc}
\toprule
\textbf{OOD Dataset} & \textbf{Clean} & \textbf{ID$\rightarrow$OOD} & \textbf{OOD$\rightarrow$ID} & \textbf{Both} \\
\midrule
CIFAR-100 & 85.71 & 77.41 & 78.39 & 68.56 \\
SVHN & 95.03 & 88.91 & 89.70 & 80.80 \\
\bottomrule \vspace{-1.5em}
\end{tabular}
\label{tab:square}
\end{table}
To complement our transfer attack analysis, we further evaluate HALO against a modified version of Square Attack \cite{andriushchenko2020square}, a gradient-free attack methodology. We change the method so that new perturbations are only added to the image if they either increase or decrease entropy for ID$\rightarrow$OOD and OOD$\rightarrow$ID attacks, respectively. We evaluated these attacks with 1000 iterations, using CIFAR-10 as the ID dataset and CIFAR-100 or SVHN as OOD datasets. The results, presented in Table \ref{tab:square}, demonstrate that HALO maintains a similar degree of robustness under this gradient-free attack as in the transfer attack. In both of these cases, HALO exhibits stronger resistance to black box attacks compared to white-box attacks, namely the PGD attack we use throughout this work. 

\section{Experimental Details} \label{apdx:ex_details}

\subsection{Sensitivity Analysis} \label{apdx:sens_details}

In Section \ref{sec:sens} we show the results of a sensitivity analysis on the hyperparameters of our model: $\eta$, $\gamma$ and $\beta$. For this experiment we use the same training setup as described in Appendix \ref{apdx:trainingsetup} with CIFAR-10 as the ID dataset. We report averages over 3 independent runs for each hyperparameter setting. For each hyperparameter test, we vary only the hyperparameter in question and hold all others fixed at their default values: $\eta=2.0$, $\gamma=0.5$ and $\beta = 3.0$. In all of our tests we set both $\beta_1 = \beta_2 = \beta$. For AUROC metrics, we report average values over all 6 OOD datasets: CIFAR-100, TinyImageNet, MNIST, SVHN, Textures and Places365. As performing a full AutoAttack evaluation is computationally expensive, we calculate robust accuracy using AutoAttack on a 2000 sample subset of the test set, using an APGD-DLR adversary. Empirically this reduced strength test results in about a 1\% increase in robust accuracy compared to full AutoAttack. As all models are tested this way, results are comparable.

\subsection{Extra Data Study} \label{apdx:extra_data_details}

In Section \ref{sec:extra_data} we explore the performance of HALO and ATD when using an additional pseudo-labelled dataset to supplement the training dataset. Here we provide a brief overview of the method of generation of this dataset as in \citet{carmon2019unlabeled}:

To obtain additional unlabeled data similar to CIFAR-10, the 80 Million Tiny Images (80M-TI) dataset \cite{torralba200880} is chosen, from which CIFAR-10 is a manually labelled subset. As most 80M-TI images do not correspond to CIFAR-10 classes, an 11-way classifier (10 CIFAR-10 classes plus an `other' class) is trained using a Wide ResNet 28-10 model \cite{zagoruyko2016wide} on a combination of CIFAR-10 and 1 million random 80M-TI images. This classifier is then applied to the remaining 80M-TI images after near-duplicates of the CIFAR-10 test set are removed. For each CIFAR-10 class, the 50,000 highest confidence images are selected to form a 500K image dataset. For consistency, we use the dataset provided by \citet{carmon2019unlabeled} rather than reproducing the dataset ourselves.

\subsection{Transferability Study} \label{apdx:tfr_details}

For our transferability study, we want to compare HALO with a strong adversary. For this reason we choose ATD as the other method as it is the next best competitor. For CIFAR-10 we use the ATD model trained with extra data as it is closer match to HALO's performance. Each model is subjected to attacks on the ID and OOD datasets, and perturbed images are stored and fed into both the source and target model. For each model we use the adversaries described in Appendix \ref{apdx:detection_attacks}. In the case of HALO, OOD detection is done using the baseline method of MSP and we use the built-in discriminator score for ATD.

\subsection{Post-processor Study} \label{apdx:pp_details}

Here we describe our post-processor study (Section \ref{sec:ppstudy}) in detail. We consider a number of methods to pair with HALO, each of which are briefly described here:
\begin{itemize}
    \item \textbf{Relative Mahalanobis Distance (RMDS)} This method works by calculating the Mahalanobis distance (MD) in the usual class-conditional fashion, but then subtracting the MD to the unconditioned mean training activations. 
    
    \item \textbf{Simplified Hopfield Energy (SHE)}: This method uses the energy function of a simplified Hopfield network as an OOD score.

    \item \textbf{Maximum Softmax Probability (MSP)}: A simple baseline that uses the maximum value of the softmax output as an OOD score.
    
    \item \textbf{ODIN}: A method that combines temperature scaling of the softmax output and small input perturbations to improve out-of-distribution detection relative to MSP.
    
    \item \textbf{OpenMax}: An approach that extends the softmax layer to estimate the probability of an input being from an unknown class. 
    
    \item \textbf{Energy Based OOD Detection (Energy)}: This method uses the energy score which is a scalar derived from the model's logits as the OOD score. 

    \item \textbf{Generalised Entropy (GEN)}: Uses a generalised form of entropy that is parameterised by two variables, $\gamma$ and $M$, as the OOD score. 
\end{itemize}

\subsection{Compute Details}

HALO models are trained on a single NVIDIA H100 GPU. Training the ResNet-18 model takes approximately 1 hour and 15 minutes. Training the PreAct ResNet-18 with extra data takes approximately 23 hours.  

\section{Additional Results} \label{apdx:extra_results}

\subsection{Ablation Study on CIFAR-10} \label{sec:ablation_study}
\begin{table*}[t!]
\small
\centering
\caption{Ablation study comparing HALO to possible alternatives. Mean values are reported over 3 independent runs with standard deviations in brackets and best values in \textbf{bold}. ID accuracy is measured on the CIFAR-10 test set. OOD AUROC is averaged over the six OOD datasets in OpenOOD \cite{yang2022openood}. All attacks are performed with $\epsilon=8/255$. Classification attacks are performed using AutoAttack \cite{croce2020reliable} and detection attacks are performed using PGD-40.}
\begin{tabular}{lcccccc}
\toprule
\multirow{2}{*}{\textbf{Model}} & \multicolumn{2}{c}{\textbf{Accuracy}} & \multicolumn{4}{c}{\textbf{Avg. AUROC}} \\
\cmidrule(lr){2-3} \cmidrule(lr){4-7}
 & \textbf{Clean} & \textbf{Robust} & \textbf{Clean} & \textbf{ID→OOD} & \textbf{OOD→ID} & \textbf{Both} \\
\midrule
TRADES & 80.57 (0.49) & \textbf{47.98} (0.31) & 71.94 (0.57) & 57.61 (0.46) & 60.76 (0.51) & 46.26 (0.36) \\
HAT & 83.64 (0.61) & 47.62 (0.48) & 76.34 (0.42) & 58.17 (1.34) & 62.58 (0.45) & 43.73 (2.21) \\
HALO w/o HL & 83.40 (0.19) & 47.10 (0.49) & 90.21 (0.75) & 75.73 (1.04) & 77.88 (1.02) & 58.83 (1.21) \\
HALO & \textbf{84.22} (0.49) & 47.55 (0.35) & \textbf{92.04} (0.23) & \textbf{77.62} (0.19) & \textbf{80.17} (0.27) & \textbf{60.41} (0.31) \\
HALO + HAT OE & 83.85 (0.25) & 42.84 (0.35) & 90.89 (0.98) & 70.50 (2.37) & 76.43 (0.82) & 51.13 (2.00) \\
\bottomrule \vspace{-0.5em}
\end{tabular}
\label{tab:ablation_results}
\end{table*}
We perform an ablation study to measure and confirm the contribution of each component part of our method. For these experiments, we aim to choose hyperparameters that produce models with comparable robust accuracy (around 47.6 in our tests). This allows us to test the \textit{comparative} advantage of each method while holding an important performance metric fixed. Table \ref{tab:ablhyper} shows the settings of hyperparameters we use for each method.

\begin{table}[htb!]
\small
\centering
\caption{Comparison of model hyperparameters for the CIFAR-10 ablation study.}
\begin{tabular}{lccc}
\toprule
\textbf{Model} & \textbf{$\eta$} & \textbf{$\beta$} & \textbf{$\gamma$} \\
\midrule
TRADES & - & 6.0 & - \\
HAT & - & 3.0 & 0.5 \\
HALO w/o HL & 2.0 & 6.0 & - \\
HALO & 2.0 & 3.0 & 0.5 \\
HALO + HAT OE & 2.0 & 3.0 & 0.5 \\
\bottomrule
\end{tabular}
\label{tab:ablhyper}
\end{table}

Table \ref{tab:ablation_results} shows the results of training with different objective functions.\footnote{We use the same hyperparameters for HALO and the HALO + HAT OE model, which results in a lower robust classification accuracy, making comparison difficult. Despite this, it is clear that the additional term degrades the model's performance significantly.} Classification attacks are performed using the standard AutoAttack adversary as implemented in \texttt{robustbench} \cite{croce2020reliable}. Detection attacks are performed using PGD-40. All attacks use $\epsilon=8/255$. We report mean values over 3 independent runs and include standard deviations in brackets.
We observe that training without an OE loss term (as in the TRADES and HAT models), while producing reasonable clean and robust accuracy, fails to produce strong detection results. Training using the HALO method but without the helper loss (HALO w/o HL) produces better detection performance than models without OE terms, but is a worse model in all respects compared to HALO with the helper loss. This verifies the importance of the helper term in our formulation.
We also study the inclusion of an additional OE loss term:
\begin{equation}
\ell_{\text{HAT OE}} = \text{KL}(f_\theta(\Tilde{\bm{x}}) , f_{\text{std}}(\bm{x} + \bm{\delta}))
\label{eq:l_hoe}
\end{equation}
where $f_{\text{std}}$ is a non-robust helper model and $\Tilde{\bm{x}} = \bm{x} + 2\bm{\delta}$ is a `helper' OE sample. This term is weighted by $\eta\cdot\gamma$ in our formulation. The introduction of this term is motivated by the improved clean/robust trade-off observed when the ID helper loss term is introduced. It is hypothesised that the inclusion of the helper loss term helps reduce the excessive margin in the decision boundary, leading to better classification performance \cite{rade2022reducing}.
Given we know that there is a correlation between classification and OOD detection performance, it is natural to wonder whether a similar term may be useful for our adversarial OOD detection performance. Empirically, we observe that training with this term worsens performance across the board, so it is not included in the proposed method. 

We hypothesise that this is due to differences in the similarity of train and test time samples for ID and OOD data. For ID data, the training set usually closely matches the test set in terms of distribution. In contrast, our training time OE dataset is not necessarily distributionally similar to test time OOD data. As such, the OE helper term's effect of decreasing the extent of adversarial robustness may be at odds with the model's ability to generalise from OE data, reducing its overall effectiveness.

\subsection{CIFAR-100 Experiments} \label{apdx:cifar100}

\begin{figure}[h!]
\centering
\includegraphics[width=0.45\textwidth]{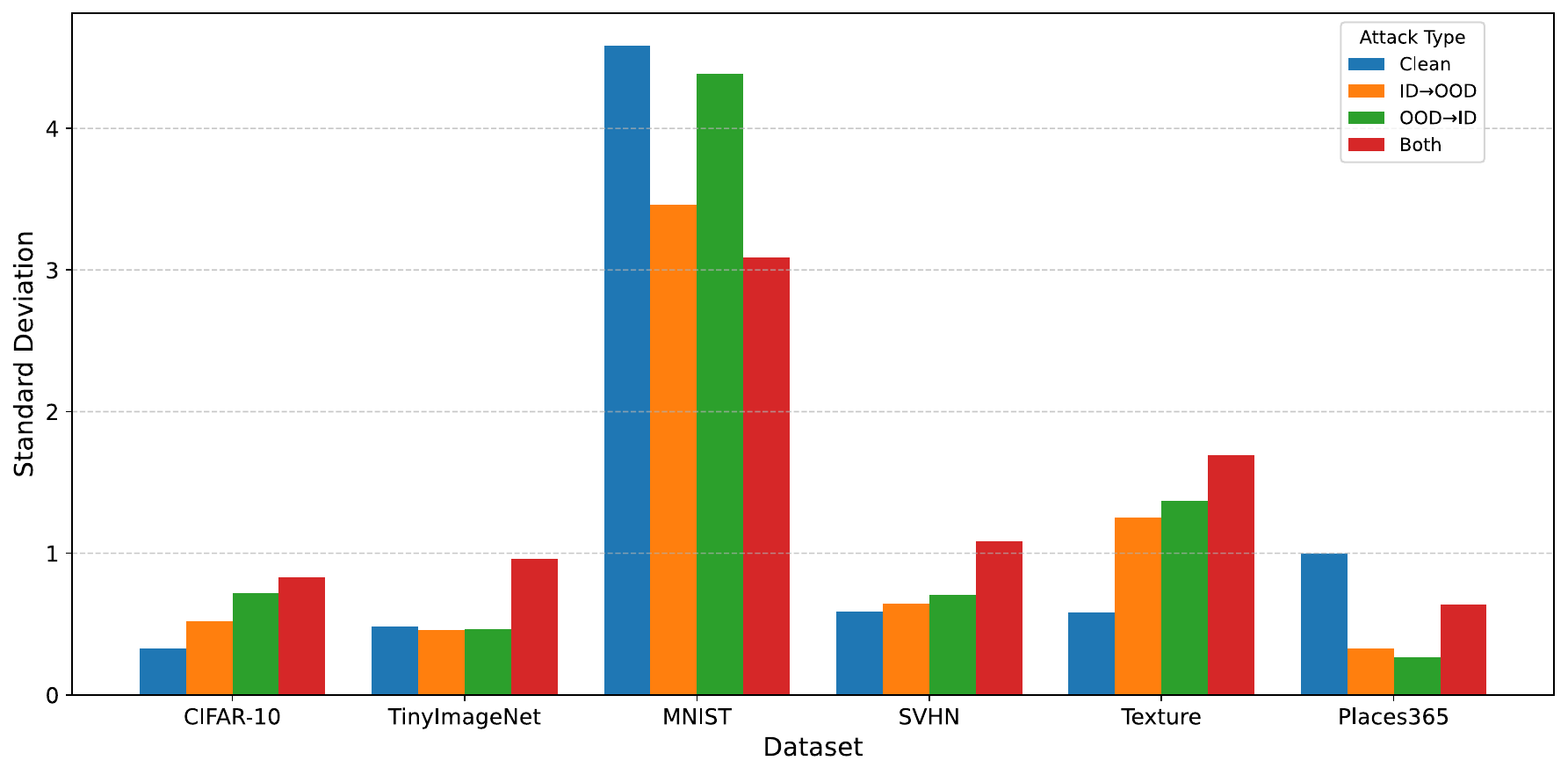}
\caption{Standard deviation of HALO's AUROC under various attack settings and datasets. CIFAR-100 is used as the ID dataset, and we use the default $\eta=2.0$, $\gamma=0.5$ and $\beta = 3.0$. MNIST has a much higher degree of variation run to run than the other datasets.}
\label{fig:mnist_std}
\end{figure}

In this section, we give results where CIFAR-100 is used as the ID dataset. For these experiments, we use the same training method as detailed in Appendix \ref{apdx:trainingsetup}. For evaluation, we use the following OOD datasets: CIFAR-10, TinyImageNet, SVHN, Texture and Places365. For CIFAR100 experiments, we exclude MNIST as we observe that results on this dataset exhibit high variance between training runs, which makes accurate measurement of performance differences difficult. We give a visualisation of this high variance in Figure \ref{fig:mnist_std}, where we show the standard deviation of HALO's AUROC scores across 3 random seeds for different attack settings. We note also that for the ablation study and sensitivity analysis, robust accuracies are calculated using the less compute intensive version of AutoAttack described in Section \ref{sec:sens}.

\textbf{Main Comparison:} In Table \ref{tab:cifar100}, we show a comparison between methods with CIFAR-100 as the ID dataset. We can see that HALO is the strongest performer among robust methods on clean and single-sided attack settings for most datasets, but is occasionally beaten in the `both attacks' setting. We note in interpreting these results that an AUROC below 50 is of little practical benefit, as it indicates worse than random performance. As an illustrative example, a hypothetical OOD detector that simply flags every input as OOD would achieve an AUROC of 50 in all of the settings we study, beating many methods we consider. Clearly, the task of robust OOD detection on CIFAR-100 is still far from solved. 

\textbf{Ablation Study:} In Table \ref{tab:cifar100_abl} we show the results of an ablation study performed with CIFAR-100 as the ID dataset. As with CIFAR-10, we aim to produce models with comparable robust accuracy. To this end, we choose the hyperparameter settings described in Table \ref{tab:ablhypercifar100}. The results of the study are similar to the CIFAR-10 case, showing that for a fixed level of clean/robust accuracy, HALO exhibits stronger robust detection than other methods. The addition of a helper OE term appears to boost clean accuracy at the cost of robust accuracy and OOD detection. indicating that it changes the nature of the robustness learned.

\begin{table*}[t!]
\centering
\small
\caption{Ablation study with CIFAR100 as the ID dataset. Mean values over 3 runs are reported with standard deviations in brackets. Best values are in \textbf{bold}. ID accuracy is measured on the CIFAR-100 test set. OOD AUROC is averaged over OOD datasets. All attacks are performed with $\epsilon=8/255$. Classification attacks use AutoAttack and detection attacks use PGD-40.}
\begin{tabular}{lcccccc}
\toprule
\multirow{2}{*}{\textbf{Model}} & \multicolumn{2}{c}{\textbf{Accuracy}} & \multicolumn{4}{c}{\textbf{Avg. AUROC}} \\
\cmidrule(lr){2-3} \cmidrule(lr){4-7}
 & \textbf{Clean} & \textbf{Robust} & \textbf{Clean} & \textbf{ID→OOD} & \textbf{OOD→ID} & \textbf{Both} \\
\midrule
TRADES & 57.99 (0.41) & 22.80 (0.88) & 69.52 (0.80) & 47.62 (0.54) & 48.68 (0.72) & 30.04 (0.46) \\
HAT & 58.29 (0.67) & \textbf{23.25} (0.74) & 69.72 (1.10) & 47.14 (0.66) & 48.31 (0.77) & 29.16 (0.20) \\
HALO w/o HL & 60.44 (0.23) & 22.88 (0.44) & 77.22 (0.53) & 49.60 (0.59) & 49.37 (0.61) & 26.71 (0.47) \\
HALO & 60.54 (0.22) & 22.78 (0.10) & \textbf{77.24} (0.33) & \textbf{52.02} (0.36) & \textbf{52.02} (0.53) & \textbf{30.05} (0.46) \\
HALO + HAT OE & \textbf{62.14} (0.45) & 21.48 (0.71) & 75.69 (0.51) & 47.39 (0.82) & 49.22 (1.05) & 26.29 (0.77) \\
\bottomrule
\end{tabular}
\label{tab:cifar100_abl}
\end{table*}

\begin{table}[h!]
\centering
\small
\caption{Comparison of model hyperparameters for the CIFAR-100 ablation study.}
\begin{tabular}{lccc}
\toprule
\textbf{Model} & \textbf{$\eta$} & \textbf{$\beta$} & \textbf{$\gamma$} \\
\midrule
TRADES & - & 3.0 & - \\
HAT & - & 2.5 & 0.2 \\
HALO w/o HL & 2.0 & 3.0 & - \\
HALO & 2.0 & 2.5 & 0.2 \\
HALO + HAT OE & 2.0 & 2.5 & 0.2 \\
\bottomrule
\end{tabular}
\label{tab:ablhypercifar100}
\end{table}

\begin{figure*}[h!]
\centering
\includegraphics[width=0.62\textwidth]{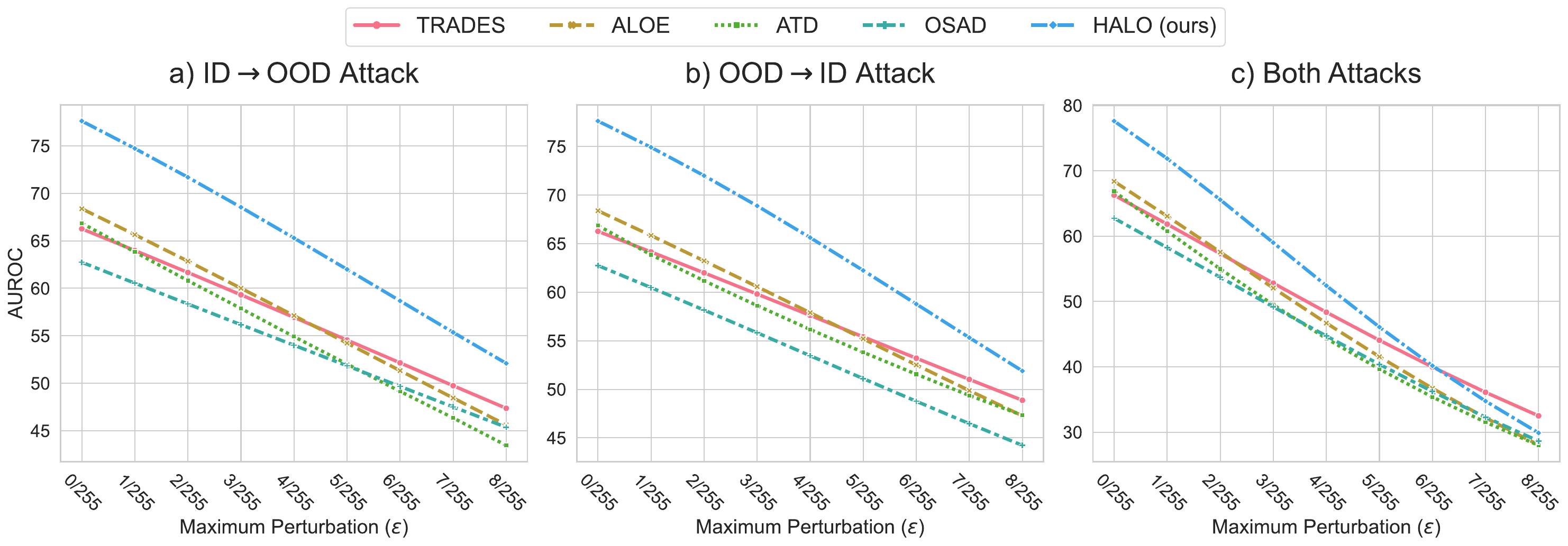}
\caption{Comparison of robust OOD detection methods across multiple attack strengths where CIFAR100 is the ID dataset. Subplots left to right: a) Only ID data is attacked, b) Only OOD data is attacked, c) Both ID and OOD data attacked. Our method, HALO, outperforms existing approaches for all reasonable attack strengths, across all threat models with the exception of high strength attacks on both ID and OOD data.}
\label{fig:eps_plot_cifar100_nm}
\end{figure*}

\textbf{Different Attack Strengths:}
In Figure \ref{fig:eps_plot_cifar100_nm} we show a comparison of different methods across a range of maximum perturbations. ATOM is excluded from this experiment as it does not achieve strong robustness and makes the plots harder to read.

\begin{figure*}[htb!]
\centering
\includegraphics[width=0.75\textwidth]{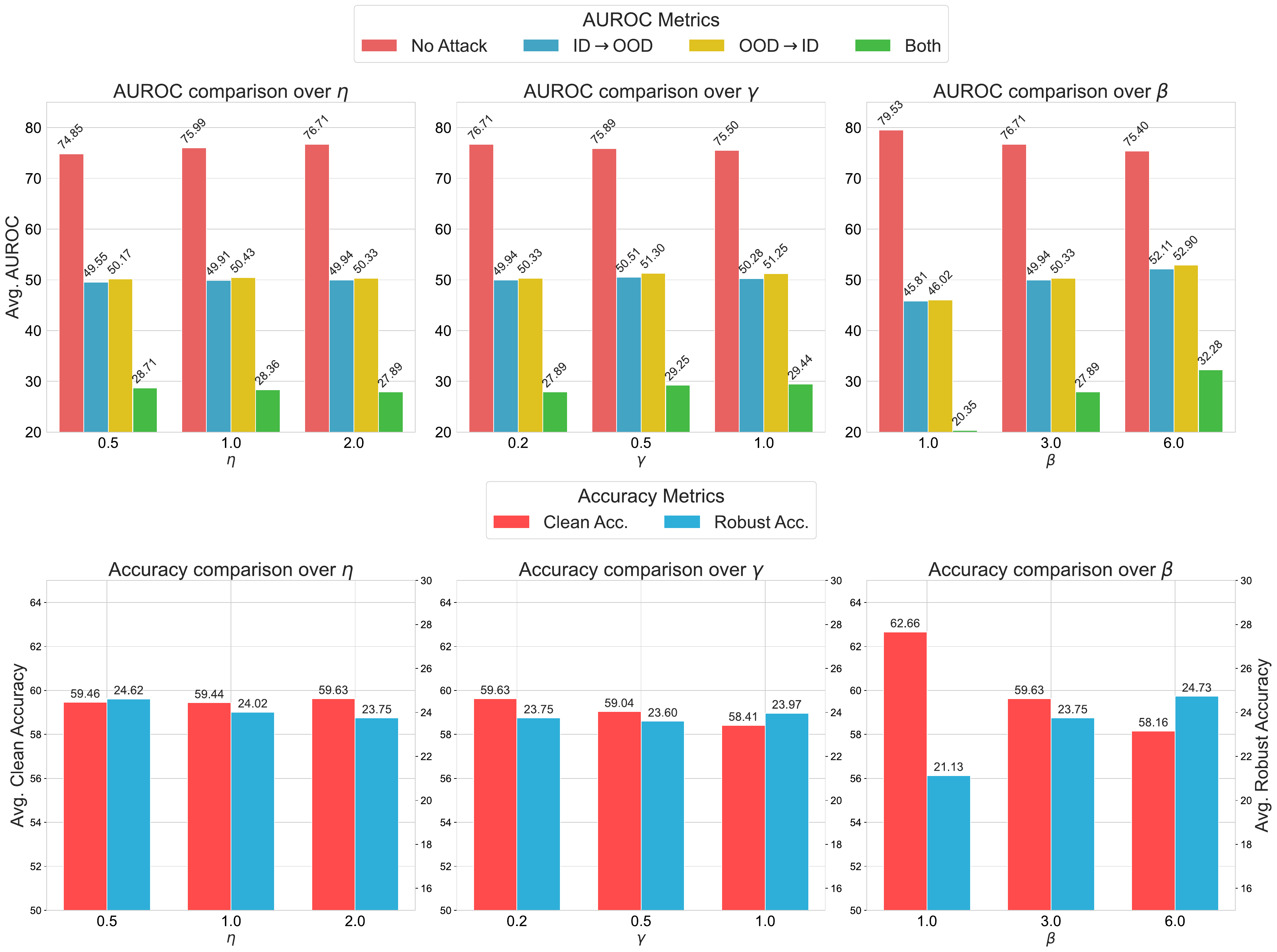}
\caption{Hyperparameter sensitivity analysis on CIFAR-100. By default, we use $\eta=2.0$, $\gamma=0.2$ and $\beta = 3.0$ and only change the hyperparameter being examined. All results are averages over 3 independent runs. Top row: averaged AUROC across 5 datasets for different attack settings, bottom row: clean accuracy (left axis) and robust accuracy (right axis). Columns: varying $\eta$ (left), $\gamma$ (middle) and $\beta$ (right).}
\label{fig:sens_cifar100}
\end{figure*}

\textbf{Sensitivity Analysis:} Figure \ref{fig:sens_cifar100} shows the results of a sensitivity analysis conducted on CIFAR-100. The trends observed in this plot are mostly similar to those observed with CIFAR-10. One notable difference is that we choose $\gamma=0.2$ as the default setting for the helper loss weighting. This is done because we want to observe the effect of changing a single hyperparameter where the others are held fixed at their optimal values.

\subsection{Results on TinyImageNet}

To test the applicability of HALO to larger, more complicated datasets, we evaluate it and a number of other methods with TinyImageNet \cite{le2015tiny} as the ID dataset. For these experiments we follow the training setup described in Appendix \ref{apdx:trainingsetup} but change the maximum learning rate to 0.05 and use a PreAct ResNet-18 as the base model. Additionally, the images are resized to 32x32. As this is an inherently more difficult problem, we use a maximum perturbation of $\epsilon=2/255$ in both training and evaluation. To evaluate OOD detection we use the following datasets: CIFAR-10, CIFAR-100, SVHN, Texture and Places365. Methods that use a post-processor are paired with MSP to produce an OOD score. Table \ref{tab:results_TI} shows the detection performance of a number of robust OOD detection methods in this setting. On most datasets, HALO has the strongest performance in clean and single-sided attack settings. As in the case of CIFAR-100, we note that robust OOD detection on TinyImageNet remains a difficult challenge, even with reduced perturbation strength. 

\subsection{Results on GTSRB}

In order to study the performance of various methods in settings closer to the real-world applications they may be deployed in, we evaluate HALO and a number of other robust OOD detection methods when the German Traffic Sign Recognition Benchmark (GTSRB) dataset \cite{stallkamp2011german} is ID. This allows us to study how these methods may perform in the real-world setting of an autonomous driving system, where the model must correctly classify street signs and reject anomalous inputs such as when fog obscures the road markings. To simulate these conditions, we use the Texture dataset as OOD input. We resize all images to 32x32, use an attack with $\epsilon=8/255$ and 10 / 40 steps for training / testing respectively and utilise the MSP post-processor for all models apart from ATD, where we use the built-in discriminator. 
In Table \ref{tab:GTSRB}, we show the results of this experiment. We see that most methods achieve strong clean performance, but falter under adversarial attacks. HALO has a significant lead over the other methods, exhibiting both strong clean performance and resistance to attacks, suggesting it as a strong choice for real-world applications.
\begin{table}[!htbp]
\small
\centering
\caption{GTSRB performance (AUROC) across various attack settings and method. Best results are bolded.}
\begin{tabular}{lccccc}
\hline
\textbf{Metric} & \textbf{OE} & \textbf{TRADES} & \textbf{ALOE} & \textbf{ATD} & \textbf{HALO} \\
\hline
\text{Clean} & 98.76 & 93.79 & 91.50 & 97.93 & \textbf{99.75} \\
\text{ID$\rightarrow$OOD} & 4.75 & 82.74 & 77.61 & 92.23 & \textbf{98.32} \\
\text{OOD$\rightarrow$ID} & 14.95 & 76.21 & 73.47 & 92.54 & \textbf{98.33} \\
\text{Both} & 0.41 & 59.52 & 53.90 & 81.00 & \textbf{94.25} \\
\hline
\end{tabular}
\label{tab:GTSRB}
\end{table}

\subsection{Class Cardinality Experiment} \label{apdx:class_cardinality}

To investigate potential factors affecting model performance on larger datasets, we conduct a systematic study examining the role of class cardinality. We construct a series of restricted CIFAR-100 datasets by selecting the first $K$ classes, where $K \in {25, 50, 75}$ and the 100 (unmodified), maintaining all images from the selected classes. This creates a controlled experimental setting where we can isolate the effect of increasing class cardinality while keeping other dataset characteristics constant.
We train HALO models on each of these datasets following the standard experimental protocol detailed in Section \ref{sec:eval}. The results of these experiments are presented in Table \ref{tab:cardinality}.
The relatively modest variations in performance across different class cardinalities suggest that this factor alone does not account for the observed performance degradation on larger datasets. This finding indicates that alternative characteristics such as dataset scale or image diversity may play more substantial roles in determining model effectiveness, though further investigation is needed to definitively identify these factors.

\subsection{Other Detection Metrics} \label{apdx:extra_det_res}

For completeness, we give our main comparative results on CIFAR-10 (as in Table \ref{tab:main-results}) in terms of FPR95 (Table \ref{tab:main-results-fpr95}) and AUPR (Table \ref{tab:main-results-auprin}).

\newpage

\setlength{\textheight}{10in} 
\setlength{\topmargin}{-0.5in}
\setlength{\headheight}{0pt}  
\setlength{\headsep}{0pt}     
\setlength{\footskip}{0in}    

\begin{figure*}[htbp]
    \begin{center}
        \caption{Comparison of AUROCs across different datasets and attack settings for each method when CIFAR-100 is the ID dataset. Methods lacking both kinds of robustness are \textbf{\textcolor{gray}{greyed out}}. Best and second-best results among robust methods are \textbf{bolded} and \underline{underlined}, respectively. The $\Delta$ column shows the difference between HALO and the best other method, colored \textcolor{green}{green} if positive and \textcolor{red}{red} if negative. All attacks are performed using PGD-40 with $\epsilon=8/255$.}
        \resizebox{0.65\textwidth}{!}{
        
            \begin{tabular}{llrrrrrrr|r}   
            \hline
            Dataset & Attack Type & \textcolor{gray}{OE} & \textcolor{gray}{ATOM} & TRADES & ALOE & OSAD & ATD & HALO & $\Delta$ \\
            \hline
            \multirow{4}{*}{CIFAR10} & Clean & \textcolor{gray}{73.63} & \textcolor{gray}{61.96} & \underline{65.31} & 64.20 & 59.68 & 56.54 & \textbf{68.17} & \textcolor{green}{+2.86} \\
            & ID$\rightarrow$OOD & \textcolor{gray}{1.85} & \textcolor{gray}{2.75} & \textbf{46.38} & 41.90 & \underline{42.61} & 33.05 & 42.54 & \textcolor{red}{-3.84} \\
            & OOD$\rightarrow$ID & \textcolor{gray}{0.00} & \textcolor{gray}{0.00} & \textbf{48.05} & \underline{43.40} & 41.89 & 38.65 & 43.04 & \textcolor{red}{-5.01} \\
            & Both & \textcolor{gray}{0.00} & \textcolor{gray}{0.00} & \textbf{31.87} & 25.35 & \underline{27.03} & 22.08 & 24.05 & \textcolor{red}{-7.82} \\
            \hline
            \multirow{4}{*}{TIN} & Clean & \textcolor{gray}{99.98} & \textcolor{gray}{47.10} & 68.89 & \underline{69.45} & 63.39 & 69.22 & \textbf{80.66} & \textcolor{green}{+11.21} \\
            & ID$\rightarrow$OOD & \textcolor{gray}{2.71} & \textcolor{gray}{1.85} & \underline{49.79} & 46.38 & 45.99 & 43.78 & \textbf{55.63} & \textcolor{green}{+5.84} \\
            & OOD$\rightarrow$ID & \textcolor{gray}{0.01} & \textcolor{gray}{0.00} & \underline{50.69} & 47.56 & 43.89 & 48.89 & \textbf{54.97} & \textcolor{green}{+4.28} \\
            & Both & \textcolor{gray}{0.00} & \textcolor{gray}{0.00} & \textbf{33.86} & 28.15 & 28.31 & 28.83 & \underline{32.08} & \textcolor{red}{-1.78} \\
            \hline
            \multirow{4}{*}{SVHN} & Clean & \textcolor{gray}{99.33} & \textcolor{gray}{80.51} & 68.94 & \underline{77.43} & 70.77 & 75.71 & \textbf{81.45} & \textcolor{green}{+4.02} \\
            & ID$\rightarrow$OOD & \textcolor{gray}{1.10} & \textcolor{gray}{4.23} & 49.04 & 53.03 & 52.43 & \underline{53.43} & \textbf{54.85} & \textcolor{green}{+1.42} \\
            & OOD$\rightarrow$ID & \textcolor{gray}{0.00} & \textcolor{gray}{0.00} & 50.12 & \textbf{54.80} & 50.63 & 53.46 & \underline{54.40} & \textcolor{red}{-0.40} \\
            & Both & \textcolor{gray}{0.00} & \textcolor{gray}{0.00} & \textbf{33.28} & \underline{33.27} & 33.23 & 31.38 & 31.37 & \textcolor{red}{-1.91} \\
            \hline
            \multirow{4}{*}{Texture} & Clean & \textcolor{gray}{86.14} & \textcolor{gray}{71.26} & 59.35 & 63.15 & 57.80 & \underline{68.85} & \textbf{79.87} & \textcolor{green}{+11.02} \\
            & ID$\rightarrow$OOD & \textcolor{gray}{0.21} & \textcolor{gray}{3.59} & 41.51 & 41.37 & 40.93 & \underline{46.40} & \textbf{55.41} & \textcolor{green}{+9.01} \\
            & OOD$\rightarrow$ID & \textcolor{gray}{0.98} & \textcolor{gray}{0.03} & 44.24 & 44.35 & 41.24 & \underline{48.87} & \textbf{56.72} & \textcolor{green}{+7.85} \\
            & Both & \textcolor{gray}{0.00} & \textcolor{gray}{0.00} & 29.09 & 26.04 & 26.58 & \underline{29.30} & \textbf{33.57} & \textcolor{green}{+4.27} \\
            \hline
            \multirow{4}{*}{Places365} & Clean & \textcolor{gray}{80.58} & \textcolor{gray}{69.40} & \underline{68.83} & 67.66 & 62.08 & 63.95 & \textbf{77.93} & \textcolor{green}{+9.10} \\
            & ID$\rightarrow$OOD & \textcolor{gray}{0.05} & \textcolor{gray}{3.39} & \underline{49.98} & 45.11 & 44.77 & 39.27 & \textbf{52.93} & \textcolor{green}{+2.95} \\
            & OOD$\rightarrow$ID & \textcolor{gray}{0.00} & \textcolor{gray}{0.00} & \underline{50.63} & 45.64 & 42.65 & 46.03 & \textbf{52.39} & \textcolor{green}{+1.76} \\
            & Both & \textcolor{gray}{0.00} & \textcolor{gray}{0.00} & \textbf{33.83} & 26.81 & 27.23 & 26.95 & \underline{30.29} & \textcolor{red}{-3.54} \\
            \hline
            \end{tabular}
        }
        \label{tab:cifar100}    
    \end{center}
\end{figure*}

\begin{table*}[t!]
\small
\centering
\caption{Comparison of AUROCs across different datasets and attack settings for each method when TinyImageNet is used as the ID dataset. All attacks are performed using PGD-40 with $\epsilon=2/255$. Methods that use a post-processor are paired with MSP. Best and second-best results are \textbf{bolded} and \underline{underlined}, respectively. The $\Delta$ column shows the difference between HALO and the best other method, colored \textcolor{green}{green} if positive and \textcolor{red}{red} if negative. }
\begin{tabular}{llrrrrr|r}   
\hline
Dataset & Attack Type & TRADES & ALOE & OSAD & ATD & HALO & $\Delta$ \\
\hline
\multirow{4}{*}{CIFAR10} & Clean & \underline{60.82} & 55.60 & 50.71 & 46.93 & \textbf{60.90} & \textcolor{green}{+0.08} \\
& ID$\rightarrow$OOD & 42.63 & 43.16 & \underline{44.05} & 32.27 & \textbf{46.78} & \textcolor{green}{+2.73} \\
& OOD$\rightarrow$ID & 41.44 & 42.60 & \underline{43.97} & 32.16 & \textbf{46.33} & \textcolor{green}{+2.36} \\
& Both & 25.86 & 31.19 & \textbf{37.52} & 20.16 & \underline{33.35} & \textcolor{red}{-4.17} \\
\hline
\multirow{4}{*}{CIFAR100} & Clean & \textbf{61.28} & 53.68 & 49.77 & 50.89 & \underline{60.87} & \textcolor{red}{-0.41} \\
& ID$\rightarrow$OOD & \underline{43.18} & 41.50 & 43.07 & 36.14 & \textbf{47.07} & \textcolor{green}{+3.89} \\
& OOD$\rightarrow$ID & \underline{43.22} & 41.34 & 42.27 & 36.77 & \textbf{46.63} & \textcolor{green}{+3.41} \\
& Both & 26.48 & 30.16 & \textbf{36.83} & 24.09 & \underline{33.73} & \textcolor{red}{-3.10} \\
\hline
\multirow{4}{*}{SVHN} & Clean & 68.51 & 57.48 & 68.65 & \textbf{78.42} & \underline{73.36} & \textcolor{red}{-5.06} \\
& ID$\rightarrow$OOD & 50.05 & 44.74 & \underline{61.82} & \textbf{63.84} & 59.67 & \textcolor{red}{-4.17} \\
& OOD$\rightarrow$ID & 48.34 & 45.37 & \underline{61.35} & \textbf{65.30} & 57.63 & \textcolor{red}{-7.67} \\
& Both & 31.24 & 33.40 & \textbf{54.28} & \underline{48.19} & 43.37 & \textcolor{red}{-10.91} \\
\hline
\multirow{4}{*}{Texture} & Clean & \underline{68.39} & 53.23 & 54.97 & 53.63 & \textbf{71.21} & \textcolor{green}{+2.82} \\
& ID$\rightarrow$OOD & \underline{49.91} & 40.55 & 48.24 & 38.90 & \textbf{57.97} & \textcolor{green}{+8.06} \\
& OOD$\rightarrow$ID & \underline{49.95} & 42.76 & 48.76 & 39.14 & \textbf{57.77} & \textcolor{green}{+7.82} \\
& Both & 32.99 & 31.04 & \underline{42.12} & 26.43 & \textbf{44.24} & \textcolor{green}{+2.12} \\
\hline
\multirow{4}{*}{Places365} & Clean & \underline{64.41} & 61.12 & 56.35 & 57.14 & \textbf{65.37} & \textcolor{green}{+0.96} \\
& ID$\rightarrow$OOD & 45.02 & 48.30 & \underline{49.51} & 41.84 & \textbf{50.45} & \textcolor{green}{+0.94} \\
& OOD$\rightarrow$ID & 43.93 & 47.26 & \underline{48.77} & 42.09 & \textbf{50.09} & \textcolor{green}{+1.32} \\
& Both & 27.35 & 35.14 & \textbf{42.05} & 28.38 & \underline{36.18} & \textcolor{red}{-5.87} \\
\hline
\end{tabular} \vspace{-0.5em}
\label{tab:results_TI}
\end{table*}

\begin{table*}[t!]
\small
\centering
\caption{Performance comparison across different class cardinalities K for HALO, showing AUROC scores for various datasets and attack settings. All experiments use CIFAR-100 as the ID dataset, with only the first K classes used.}
\begin{tabular}{llrrrrr}   
\hline
Dataset & Attack Type & K=25 & K=50 & K=75 & K=100 \\
\hline
\multirow{4}{*}{CIFAR10} & Clean & 74.09 & 69.56 & 68.81 & 68.17 \\
& ID$\rightarrow$OOD & 50.92 & 43.02 & 44.08 & 42.54 \\
& OOD$\rightarrow$ID & 50.62 & 44.09 & 44.54 & 43.04 \\
& Both & 30.68 & 24.59 & 25.40 & 24.05 \\
\hline
\multirow{4}{*}{TIN} & Clean & 82.42 & 80.99 & 80.88 & 80.66 \\
& ID$\rightarrow$OOD & 60.33 & 55.38 & 57.28 & 55.63 \\
& OOD$\rightarrow$ID & 59.88 & 55.51 & 56.64 & 54.97 \\
& Both & 37.41 & 32.02 & 33.87 & 32.08 \\
\hline
\multirow{4}{*}{SVHN} & Clean & 80.03 & 80.65 & 82.52 & 81.45 \\
& ID$\rightarrow$OOD & 56.50 & 52.89 & 58.05 & 54.85 \\
& OOD$\rightarrow$ID & 56.21 & 53.16 & 56.08 & 54.40 \\
& Both & 34.55 & 29.91 & 33.22 & 31.37 \\
\hline
\multirow{4}{*}{Texture} & Clean & 81.96 & 79.04 & 79.11 & 79.87 \\
& ID$\rightarrow$OOD & 61.81 & 54.48 & 57.06 & 55.41 \\
& OOD$\rightarrow$ID & 62.36 & 55.68 & 57.51 & 56.72 \\
& Both & 40.53 & 32.95 & 35.33 & 33.57 \\
\hline
\multirow{4}{*}{Places365} & Clean & 81.42 & 78.77 & 77.93 & 77.93 \\
& ID$\rightarrow$OOD & 59.50 & 53.21 & 54.86 & 52.93 \\
& OOD$\rightarrow$ID & 58.63 & 53.14 & 53.94 & 52.39 \\
& Both & 36.56 & 30.40 & 32.08 & 30.29 \\
\hline
\end{tabular}
\label{tab:cardinality}
\end{table*}

\begin{table*}[t!]
\small
\centering
\caption{Comparison of FPR95 values across different datasets and attack settings for each method. Lower values are better. Methods lacking both kinds of robustness are \textbf{\textcolor{gray}{greyed out}}. Best and second-best results among robust methods are \textbf{bolded} and \underline{underlined}, respectively. The $\Delta$ column shows the difference between HALO and the best other method, colored \textcolor{green}{green} if negative and \textcolor{red}{red} if positive. CIFAR-10 is used as the ID dataset.}
\begin{tabular}{llrrrrrrr|r}   
\hline
Dataset & Attack Type & \textcolor{gray}{OE} & \textcolor{gray}{ATOM} & TRADES & ALOE & OSAD & ATD & HALO & $\Delta$ \\
\hline
\multirow{4}{*}{CIFAR100} & Clean & \textcolor{gray}{36.92} & \textcolor{gray}{39.44} & 72.39 & 67.34 & 62.28 & \underline{60.00} & \textbf{57.33} & \textcolor{green}{-2.67} \\
 & ID$\rightarrow$OOD & \textcolor{gray}{100.00} & \textcolor{gray}{100.00} & 81.60 & 83.08 & 80.20 & \underline{79.74} & \textbf{72.90} & \textcolor{green}{-6.84} \\
 & OOD$\rightarrow$ID & \textcolor{gray}{100.00} & \textcolor{gray}{91.79} & 82.36 & 80.11 & 77.11 & \underline{76.64} & \textbf{73.22} & \textcolor{green}{-3.42} \\
 & Both & \textcolor{gray}{100.00} & \textcolor{gray}{100.00} & 88.81 & 91.22 & 90.47 & \underline{88.94} & \textbf{84.03} & \textcolor{green}{-4.91} \\
\hline
\multirow{4}{*}{TIN} & Clean & \textcolor{gray}{0.04} & \textcolor{gray}{13.52} & 62.12 & 61.24 & 52.73 & \underline{47.64} & \textbf{37.27} & \textcolor{green}{-10.37} \\
 & ID$\rightarrow$OOD & \textcolor{gray}{95.50} & \textcolor{gray}{99.90} & 73.66 & 78.71 & 73.77 & \underline{73.74} & \textbf{58.62} & \textcolor{green}{-15.12} \\
 & OOD$\rightarrow$ID & \textcolor{gray}{100.00} & \textcolor{gray}{77.42} & 74.07 & 75.34 & \underline{69.26} & 69.91 & \textbf{59.50} & \textcolor{green}{-9.76} \\
 & Both & \textcolor{gray}{100.00} & \textcolor{gray}{100.00} & 83.06 & 88.41 & \underline{85.09} & 85.12 & \textbf{74.62} & \textcolor{green}{-10.47} \\
\hline
\multirow{4}{*}{SVHN} & Clean & \textcolor{gray}{1.18} & \textcolor{gray}{5.58} & 51.86 & 48.00 & 38.32 & \underline{34.70} & \textbf{18.81} & \textcolor{green}{-15.89} \\
 & ID$\rightarrow$OOD & \textcolor{gray}{100.00} & \textcolor{gray}{99.77} & 65.33 & 70.56 & \underline{61.98} & 67.26 & \textbf{42.82} & \textcolor{green}{-19.16} \\
 & OOD$\rightarrow$ID & \textcolor{gray}{100.00} & \textcolor{gray}{88.03} & 68.93 & 67.57 & \underline{56.06} & 56.88 & \textbf{43.31} & \textcolor{green}{-12.75} \\
 & Both & \textcolor{gray}{100.00} & \textcolor{gray}{100.00} & 78.99 & 83.27 & \underline{76.17} & 78.24 & \textbf{62.60} & \textcolor{green}{-13.57} \\
\hline
\multirow{4}{*}{Texture} & Clean & \textcolor{gray}{18.06} & \textcolor{gray}{5.13} & 64.03 & 58.61 & 45.37 & \underline{32.57} & \textbf{16.36} & \textcolor{green}{-16.21} \\
 & ID$\rightarrow$OOD & \textcolor{gray}{100.00} & \textcolor{gray}{99.76} & 75.16 & 76.87 & 67.71 & \underline{66.17} & \textbf{39.90} & \textcolor{green}{-26.27} \\
 & OOD$\rightarrow$ID & \textcolor{gray}{100.00} & \textcolor{gray}{63.88} & 74.68 & 71.79 & 60.21 & \underline{54.97} & \textbf{36.99} & \textcolor{green}{-17.98} \\
 & Both & \textcolor{gray}{100.00} & \textcolor{gray}{100.00} & 83.40 & 86.30 & 78.79 & \underline{77.36} & \textbf{58.36} & \textcolor{green}{-19.00} \\
\hline
\multirow{4}{*}{Places365} & Clean & \textcolor{gray}{28.07} & \textcolor{gray}{16.38} & 63.18 & 64.21 & 53.33 & \underline{52.77} & \textbf{41.99} & \textcolor{green}{-10.78} \\
 & ID$\rightarrow$OOD & \textcolor{gray}{100.00} & \textcolor{gray}{99.96} & \underline{74.44} & 80.72 & 74.46 & 76.46 & \textbf{61.63} & \textcolor{green}{-12.81} \\
 & OOD$\rightarrow$ID & \textcolor{gray}{100.00} & \textcolor{gray}{79.03} & 74.82 & 77.06 & \underline{71.73} & 72.91 & \textbf{62.11} & \textcolor{green}{-9.62} \\
 & Both & \textcolor{gray}{100.00} & \textcolor{gray}{100.00} & \underline{83.50} & 89.46 & 86.80 & 86.77 & \textbf{76.30} & \textcolor{green}{-7.20} \\
\hline
\multirow{4}{*}{MNIST} & Clean & \textcolor{gray}{2.42} & \textcolor{gray}{12.57} & 88.38 & 68.07 & 71.72 & \textbf{0.97} & \underline{43.04} & \textcolor{red}{+42.07} \\
 & ID$\rightarrow$OOD & \textcolor{gray}{100.00} & \textcolor{gray}{99.90} & 93.11 & 83.66 & 86.77 & \textbf{7.47} & \underline{62.44} & \textcolor{red}{+54.97} \\
 & OOD$\rightarrow$ID & \textcolor{gray}{100.00} & \textcolor{gray}{49.11} & 90.98 & 72.62 & 75.89 & \textbf{1.28} & \underline{52.29} & \textcolor{red}{+51.01} \\
 & Both & \textcolor{gray}{100.00} & \textcolor{gray}{100.00} & 94.82 & 86.82 & 89.79 & \textbf{9.89} & \underline{69.26} & \textcolor{red}{+59.37} \\
\hline
\end{tabular}
\label{tab:main-results-fpr95}
\end{table*}

\begin{table*}[t!]
\small
\centering
\caption{Comparison of AUPR values across different datasets and attack settings for each method. Higher values are better. Methods lacking both kinds of robustness are \textbf{\textcolor{gray}{greyed out}}. Best and second-best results among robust methods are \textbf{bolded} and \underline{underlined}, respectively. The $\Delta$ column shows the difference between HALO and the best other method, colored \textcolor{green}{green} if positive and \textcolor{red}{red} if negative. CIFAR-10 is used as the ID dataset.}
\begin{tabular}{llrrrrrrr|r}   
\hline
Dataset & Attack Type & \textcolor{gray}{OE} & \textcolor{gray}{ATOM} & TRADES & ALOE & OSAD & ATD & HALO & $\Delta$ \\
\hline
\multirow{4}{*}{CIFAR100} & Clean & \textcolor{gray}{91.27} & \textcolor{gray}{91.40} & 73.86 & 76.29 & 77.59 & \underline{81.09} & \textbf{83.45} & \textcolor{green}{+2.36} \\
 & ID$\rightarrow$OOD & \textcolor{gray}{30.35} & \textcolor{gray}{30.95} & \underline{63.08} & 58.63 & 60.02 & 60.37 & \textbf{70.40} & \textcolor{green}{+7.32} \\
 & OOD$\rightarrow$ID & \textcolor{gray}{30.72} & \textcolor{gray}{52.66} & 63.01 & 61.54 & 63.35 & \underline{65.27} & \textbf{70.57} & \textcolor{green}{+5.30} \\
 & Both & \textcolor{gray}{30.34} & \textcolor{gray}{30.67} & \underline{52.72} & 45.55 & 46.50 & 46.99 & \textbf{56.84} & \textcolor{green}{+4.12} \\
\hline
\multirow{4}{*}{TIN} & Clean & \textcolor{gray}{99.97} & \textcolor{gray}{97.76} & 81.40 & 82.87 & 85.31 & \underline{89.20} & \textbf{92.33} & \textcolor{green}{+3.13} \\
 & ID$\rightarrow$OOD & \textcolor{gray}{35.65} & \textcolor{gray}{34.21} & \underline{71.41} & 66.47 & 69.67 & 70.36 & \textbf{81.98} & \textcolor{green}{+10.57} \\
 & OOD$\rightarrow$ID & \textcolor{gray}{33.61} & \textcolor{gray}{69.40} & 71.99 & 69.35 & 72.18 & \underline{75.70} & \textbf{82.06} & \textcolor{green}{+6.36} \\
 & Both & \textcolor{gray}{33.13} & \textcolor{gray}{33.50} & \underline{61.63} & 52.97 & 55.49 & 56.44 & \textbf{68.76} & \textcolor{green}{+7.13} \\
\hline
\multirow{4}{*}{SVHN} & Clean & \textcolor{gray}{99.25} & \textcolor{gray}{97.03} & 69.70 & 72.19 & 79.95 & \underline{81.22} & \textbf{90.90} & \textcolor{green}{+9.68} \\
 & ID$\rightarrow$OOD & \textcolor{gray}{14.13} & \textcolor{gray}{16.77} & 55.04 & 48.52 & \underline{57.26} & 51.64 & \textbf{74.65} & \textcolor{green}{+17.39} \\
 & OOD$\rightarrow$ID & \textcolor{gray}{14.11} & \textcolor{gray}{32.11} & 53.51 & 52.42 & \underline{63.11} & 61.36 & \textbf{75.21} & \textcolor{green}{+12.10} \\
 & Both & \textcolor{gray}{14.02} & \textcolor{gray}{14.11} & 40.23 & 32.94 & \underline{41.03} & 37.46 & \textbf{55.70} & \textcolor{green}{+14.67} \\
\hline
\multirow{4}{*}{Texture} & Clean & \textcolor{gray}{97.82} & \textcolor{gray}{99.30} & 83.69 & 86.11 & 90.04 & \underline{94.18} & \textbf{97.75} & \textcolor{green}{+3.57} \\
 & ID$\rightarrow$OOD & \textcolor{gray}{39.87} & \textcolor{gray}{42.17} & 74.99 & 71.79 & 76.97 & \underline{79.99} & \textbf{91.81} & \textcolor{green}{+11.82} \\
 & OOD$\rightarrow$ID & \textcolor{gray}{41.28} & \textcolor{gray}{84.87} & 77.01 & 76.03 & 82.03 & \underline{84.44} & \textbf{92.82} & \textcolor{green}{+8.38} \\
 & Both & \textcolor{gray}{39.72} & \textcolor{gray}{40.25} & \underline{67.91} & 61.33 & 67.48 & 67.61 & \textbf{83.19} & \textcolor{green}{+15.28} \\
\hline
\multirow{4}{*}{Places365} & Clean & \textcolor{gray}{82.88} & \textcolor{gray}{91.18} & 56.59 & 58.75 & 65.82 & \underline{69.34} & \textbf{76.07} & \textcolor{green}{+6.73} \\
 & ID$\rightarrow$OOD & \textcolor{gray}{10.90} & \textcolor{gray}{11.12} & 42.34 & 36.17 & \underline{42.35} & 41.15 & \textbf{56.78} & \textcolor{green}{+14.43} \\
 & OOD$\rightarrow$ID & \textcolor{gray}{10.96} & \textcolor{gray}{36.43} & 41.71 & 39.58 & 44.32 & \underline{45.79} & \textbf{56.15} & \textcolor{green}{+10.36} \\
 & Both & \textcolor{gray}{10.90} & \textcolor{gray}{10.95} & \underline{29.82} & 22.63 & 24.84 & 25.72 & \textbf{38.12} & \textcolor{green}{+8.30} \\
\hline
\multirow{4}{*}{MNIST} & Clean & \textcolor{gray}{97.25} & \textcolor{gray}{86.08} & 17.21 & 40.24 & 35.60 & \textbf{98.91} & \underline{65.65} & \textcolor{red}{-33.26} \\
 & ID$\rightarrow$OOD & \textcolor{gray}{5.93} & \textcolor{gray}{5.99} & 11.80 & 20.81 & 17.73 & \textbf{92.70} & \underline{46.17} & \textcolor{red}{-46.53} \\
 & OOD$\rightarrow$ID & \textcolor{gray}{5.93} & \textcolor{gray}{57.56} & 14.39 & 34.45 & 29.95 & \textbf{98.09} & \underline{56.87} & \textcolor{red}{-41.22} \\
 & Both & \textcolor{gray}{5.91} & \textcolor{gray}{5.94} & 10.25 & 17.38 & 14.24 & \textbf{89.18} & \underline{38.05} & \textcolor{red}{-51.13} \\
\hline
\end{tabular}
\label{tab:main-results-auprin}
\end{table*}
\end{document}